\pdfoutput=1
\documentclass{article}
\usepackage{iclr2027_conference,times}

\usepackage{amsmath,amsfonts,bm}

\def\eqref#1{equation~\ref{#1}}

\def\1{\bm{1}}

\DeclareMathAlphabet{\mathsfit}{\encodingdefault}{\sfdefault}{m}{sl}
\SetMathAlphabet{\mathsfit}{bold}{\encodingdefault}{\sfdefault}{bx}{n}

\usepackage{hyperref}
\usepackage{url}
\usepackage{cleveref}
\crefname{equation}{equation}{equations}\Crefname{equation}{Equation}{Equations}
\creflabelformat{equation}{#2#1#3}
\usepackage{enumitem}
\usepackage{amsmath,amssymb}
\usepackage{graphicx}
\usepackage{booktabs}
\usepackage{placeins}
\usepackage{xcolor}

\providecommand{\Gpre}{G_{\text{pre}}}
\providecommand{\Gft}{G_{\text{ft}}}
\providecommand{\Llast}{\mathcal{L}_{\text{last}}}
\usepackage{tikz}
\usetikzlibrary{arrows.meta,decorations.pathreplacing}
\definecolor{cA}{HTML}{2D6CDF}
\definecolor{cB}{HTML}{E03B3B}
\definecolor{cC}{HTML}{2FA84F}
\definecolor{cD}{HTML}{F29325}
\definecolor{cE}{HTML}{8E5EA2}
\definecolor{cF}{HTML}{17A2B8}

\title{Two Timescales of Forgetting in Fine-tuning of Language Models: Fast Behavioral Suppression and Slow Representational Drift}

\author{Yilin Luo$^{1,2,*}$, Jin Hwa Lee$^{1,2,*}$, Andrew M. Saxe$^{1,2}$~\& Antonio Sclocchi$^{1,3}$ \\
$^{1}$Principia \\
$^{2}$University College London \\
$^{3}$Universit\'e Paris-Saclay, LISN\\
$^{*}$Equal contribution \\
}

\iclrfinalcopy
\begin{document}

\maketitle
\lhead{Preprint}

\begin{abstract}
Today, a core pipeline of artificial intelligence systems is pretraining on a broad range of text corpora and then fine-tuning on specific tasks or desired behaviors. 
Yet we still have limited understanding of how fine-tuning on a restricted data distribution modifies representations acquired during pretraining. In particular, it remains unclear whether surface behavioral suppression happens together with the erasure of latent representations that support it. In this work, we study supervised fine-tuning (SFT) by pretraining transformers on a controlled context-free grammar with hierarchical latent structure, the Random Hierarchy Model (RHM), and fine-tuning with a restricted subset of its generative rules. 
While during pretraining the model's ability to generate with the correct grammar rules coincides with building the internal representations of the corresponding latent, we find that SFT exhibits a very different dynamics:
(i) the generation of deleted rules changes on a fast timescale inversely proportional to the magnitude of the token--latent correlation; meanwhile, their corresponding representations stay intact;
(ii) the hidden representations of the deleted rules become forgotten on a much longer timescale, associated with a slow drift of the network parameters.
Finally, we observe analogous results in a realistic setup with natural language, suggesting that this timescale separation in forgetting is a more general phenomenon of fine-tuning language models.
\end{abstract}

\section{Introduction}
\label{sec:intro}

Large language models (LLMs) are pretrained on broad text corpora and then adapted through supervised fine-tuning (SFT) for downstream tasks, instruction following, or safety alignment. Fine-tuning typically uses a narrower data distribution and can impair capabilities acquired during pretraining, a phenomenon known as \textit{forgetting}~\citep{mccloskey1989catastrophic,ramasesh2022effect,shi2025continual}. Yet a drop in an evaluation metric of the model based on its behavioral outputs does not imply that the information supporting it has been erased from its internal representations~\citep{davari2022probing,zheng2025spurious,kotha2024understanding,lynch2024eight}. Understanding this distinction is essential for determining what fine-tuning changes and what evaluations of model behavior can establish.

Despite these observations, a quantitative account of the relationship between behavioral and representational forgetting remains limited. How quickly does fine-tuning suppress previously learned behavior? How long do the underlying representations persist, and what determines their eventual change? Addressing these questions requires a setting in which the relevant latent structure is known and its role in generation can be distinguished from its use in interpreting context.

We study these questions using the Random Hierarchy Model (RHM), a probabilistic context-free grammar with a known hierarchy of latent variables~\citep{cagnetta2024:rhm}. We pretrain decoder-only transformers by next-token prediction and then fine-tune them on sequences generated after deleting a subset of the grammar's production rules. This restriction changes which sequences the model should generate without introducing new rules to learn. The known generative structure lets us separately track the use of deleted rules in model outputs and the ability to use deleted rules in context to predict allowed continuations.

Our central finding is a separation between two timescales of forgetting. Generation using deleted rules is rapidly suppressed, while the model retains representations that allow it to interpret those rules in context and use them for prediction. This representational and predictive ability deteriorates only after much longer fine-tuning. Our analysis supports an explanation in which early behavioral adaptation occurs with little change to learned features, whereas continued noisy parameter drift gradually alters the representations that support the suppressed behavior.

\paragraph{Our contributions}
\begin{itemize}
    \item We identify two timescales of forgetting in a transformer trained on hierarchically structured data: a rapid suppression of behavior and a slow drift of representations.
    \item We show that the timescale of suppression increases with the hierarchical level of the rules and with the number of rules retained. We propose a theoretical explanation of this scaling with a linear model on fixed features: for the first fine-tuning step, the gradient signal is proportional to the difference in token--latent correlations between the fine-tuning and the pretraining grammar, and the inverse magnitude of this signal is consistent with the measured timescales.
    \item We show that the forgetting of representations is caused by a drift driven by accumulated noise in the parameter updates, which continues long after behavior has been suppressed. The time to this forgetting scales with the inverse square of the learning rate, as expected for noise accumulated over the updates of Adam, whereas the time to behavior suppression scales with the inverse of the learning rate. We further show that the forgetting of representations can be advanced by weight decay, and delayed or prevented by a pull toward the pretrained parameters with explicit regularization.
    \item We observe an analogous timescale separation between behavioral and representational forgetting in a transformer pretrained on bilingual text and fine-tuned on one language, extending the phenomenon beyond the synthetic grammar setting.
\end{itemize}

These findings show why evaluating forgetting requires examining both generated outputs and latent representations of previously learned structure. They also motivate a distinction relevant to safety fine-tuning: suppressing an unwanted behavior alone does not ensure that the representations supporting it have been removed, and the two phenomena occur through distinct dynamics.

\section{Related work}
\label{sec:related}
\paragraph{Continual learning.}
A bulk of continual learning literature studied \textit{catastrophic forgetting}, where the performance of neural networks deteriorates as new tasks are learned~\citep{goodfellow2013empirical,lee2021continual,ramasesh2021anatomy,braun2022exact}. It has also been shown that large weights after the initial learning period can be one of the causes for a loss of plasticity~\citep{dohare2024loss,nikishin2022primacy,lyle2023understanding}. Yet a decrease in task accuracy does not imply a comparable loss of information from hidden representations, and substantial retrieval of the information is plausible even after the model seems to forget in its output \citep{davari2022probing,lynch2024eight,kotha2024understanding}.
In fact, selective-unlearning research has distinguished suppression of observable behavior from removal of information retained in network weights \citep{golatkar2020eternal,golatkar2020forgetting}. \citet{malladi2023kernel} characterize fine-tuning of pretrained language models with the Neural Tangent Kernel (NTK). We similarly observe that the faster timescale of behavioral change happens while latent representations stay intact. 
\citet{blanc2020implicit, li2021happens} show that noisy gradient updates near the minimizer manifold drive a slow drift, consistent with noise-induced drift of hidden representations \citep{pashakhanloo2023stochastic}. Our work connects these dynamics to forgetting in restricted fine-tuning,
where such drift erodes the representation of deleted rules after their generation has been suppressed.

\paragraph{Learnability of the Random Hierarchy Model.}
The RHM is a synthetic generative grammar whose hierarchical latent structure allows the study of representation learning and its sample complexity~\citep{cagnetta2024:rhm}. For autoregressive models, \citet{cagnetta2024:ltp1,cagnetta2025:ltp2} relate the stepwise acquisition of successive rule levels in RHM to the amount of data needed to resolve correlations over longer contexts. \citet{favero2025diffusion} study the acquisition of hierarchical compositional structure in training diffusion models, finding an analogous result. These works describe how hierarchical structure is acquired from randomly initialized networks and its corresponding sample complexities. In this work, we focus on how the previously learned hierarchical latent structure changes throughout the fine-tuning with restricted data distribution. 

\section{Random Hierarchy Model}
\label{sec:rhm-schematic}

\subsection{Pretraining and fine-tuning setup}
\label{sec:forget-setup}

\begin{figure}[t]
\centering
\begin{tikzpicture}[line width=0.5pt,
    symS/.style={circle,draw=none,fill=#1,minimum size=1.8mm,inner sep=0pt},
    symD/.style={symS=#1,opacity=0.30},
    card/.style={rounded corners=0.8mm,draw=black!25,fill=#1},
    nd/.style={circle,draw=none,fill=#1,minimum size=3.2mm,inner sep=0pt,
               text=black,font=\fontsize{5}{5.5}\selectfont},
    ndl/.style={nd={#1!55}},
    ed/.style={-{Latex[length=1.0mm,width=0.85mm]},black!55},
    eds/.style={-{Latex[length=0.8mm,width=0.7mm]},black!55,line width=0.4pt},
    edsD/.style={eds,opacity=0.30},
    lv/.style={font=\scriptsize,text=black},
    hd/.style={font=\footnotesize\bfseries},
    mn/.style={circle,draw=none,fill=black!25,minimum size=1.3mm,inner sep=0pt},
    tk/.style={circle,draw=none,fill=#1,minimum size=1.7mm,inner sep=0pt},
    emd/.style={-{Latex[length=0.8mm,width=0.7mm]},black!40,line width=0.35pt,
                shorten >=0.8mm},
    emu/.style={{Latex[length=0.8mm,width=0.7mm]}-,black!40,line width=0.35pt,
                shorten <=0.7mm},
    fbox/.style={rounded corners=1mm,draw=black!35,fill=black!3,align=center,
                 inner xsep=4pt,inner ysep=2.5pt,font=\footnotesize},
    >={Latex[length=1.5mm,width=1.2mm]}]

  \begin{scope}[xshift=0.63cm]
  \node[hd] at (1.0,3.2) {(a) The rule books};
  \foreach \k in {3,2,1}{
    \path[card=black!7] ({0.08*\k},{-0.47+0.21*\k})
            rectangle ({1.8+0.08*\k},{2.05+0.21*\k});
    \node[lv,anchor=west] at ({0.03+0.08*\k},{1.92+0.21*\k})
      {level \the\numexpr\k+1\relax};}
  \path[card=black!3] (0,-0.47) rectangle (1.8,2.05);
  \node[lv,anchor=west] at (0.03,1.92) {level 1};
  \foreach \x/\p/\q in {0.42/cB/cC, 1.38/cB/cE}{
    \node[symS=cA] (pa) at (\x,1.34) {};
    \node[symS=\p] (ca) at ({\x-0.14},0.98) {};
    \node[symS=\q] (cb) at ({\x+0.14},0.98) {};
    \draw[eds] (pa) -- (ca); \draw[eds] (pa) -- (cb);}
  \node[lv] at (0.90,1.16) {$\cdots$};
  \draw[decorate,decoration={brace,amplitude=2pt},black!55] (0.26,1.49) -- (1.54,1.49);
  \node[lv] at (0.90,1.62) {$m$};
  \draw[decorate,decoration={brace,mirror,amplitude=2pt},black!55] (0.19,0.86) -- (0.65,0.86);
  \node[lv] at (0.42,0.72) {$s$};
  \foreach \x/\p/\q in {0.42/cF/cB, 1.38/cC/cB}{
    \node[symS=cC] (pa) at (\x,0.44) {};
    \node[symS=\p] (ca) at ({\x-0.14},0.08) {};
    \node[symS=\q] (cb) at ({\x+0.14},0.08) {};
    \draw[eds] (pa) -- (ca); \draw[eds] (pa) -- (cb);}
  \node[lv] at (0.90,0.26) {$\cdots$};
  \node[lv] at (0.42,-0.16) {$\vdots$};
  \node[lv,anchor=west] at (0.62,-0.30) {$v$ symbols};

  \end{scope}

  \begin{scope}[xshift=4.40cm]
  \node[hd] at (1.0,3.2) {(b) The restriction};
  \foreach \k in {3,2,1}
    \path[card=black!7] ({0.06*\k},{-0.47+0.11*\k})
            rectangle ({1.8+0.06*\k},{2.05+0.11*\k});
  \path[card=black!3] (0,-0.47) rectangle (1.8,2.05);
  \node[lv,anchor=west] at (0.03,1.92) {every level};
  \foreach \x/\p/\q in {0.28/cB/cC, 0.69/cB/cE}{
    \node[symS=cA] (pa) at (\x,1.34) {};
    \node[symS=\p] (ca) at ({\x-0.13},0.98) {};
    \node[symS=\q] (cb) at ({\x+0.13},0.98) {};
    \draw[eds] (pa) -- (ca); \draw[eds] (pa) -- (cb);}
  \foreach \x/\p/\q in {1.10/cC/cD, 1.51/cF/cB}{
    \node[symD=cA] (pa) at (\x,1.34) {};
    \node[symD=\p] (ca) at ({\x-0.13},0.98) {};
    \node[symD=\q] (cb) at ({\x+0.13},0.98) {};
    \draw[edsD] (pa) -- (ca); \draw[edsD] (pa) -- (cb);}
  \draw[cB,line width=0.6pt] (0.92,0.98) -- (1.69,0.98);
  \draw[decorate,decoration={brace,amplitude=2pt},black!55] (0.12,1.49) -- (1.67,1.49);
  \node[lv] at (0.90,1.62) {$m$};
  \draw[decorate,decoration={brace,mirror,amplitude=2pt},black!55] (0.12,0.86) -- (0.85,0.86);
  \node[lv] at (0.48,0.72) {kept};
  \draw[decorate,decoration={brace,mirror,amplitude=2pt},cB] (0.94,0.86) -- (1.67,0.86);
  \node[lv] at (1.30,0.72) {deleted};
  \foreach \x/\p/\q in {0.28/cF/cB, 0.69/cC/cB}{
    \node[symS=cC] (pa) at (\x,0.44) {};
    \node[symS=\p] (ca) at ({\x-0.13},0.08) {};
    \node[symS=\q] (cb) at ({\x+0.13},0.08) {};
    \draw[eds] (pa) -- (ca); \draw[eds] (pa) -- (cb);}
  \foreach \x/\p/\q in {1.10/cD/cE, 1.51/cB/cF}{
    \node[symD=cC] (pa) at (\x,0.44) {};
    \node[symD=\p] (ca) at ({\x-0.13},0.08) {};
    \node[symD=\q] (cb) at ({\x+0.13},0.08) {};
    \draw[edsD] (pa) -- (ca); \draw[edsD] (pa) -- (cb);}
  \draw[cB,line width=0.6pt] (0.92,0.08) -- (1.69,0.08);
  \node[lv] at (0.28,-0.16) {$\vdots$};
  \end{scope}

  \begin{scope}[xshift=8.725cm]
  \node[hd] at (1.965,3.2) {(c) Generating a sequence};
  \foreach \y/\l in {2.6/4, 1.95/3, 1.3/2, 0.65/1, 0.0/0}
    \node[lv,anchor=east] at (-0.24,\y) {level \l};
  \node[ndl=cD] (r) at ({0.33*7.5},2.6) {$\mu^{4}_{1}$};
  \foreach \i/\c in {0/cB, 1/cE}
    \node[ndl=\c] (a\i) at ({0.33*(3.5+8*\i)},1.95) {$\mu^{3}_{\the\numexpr\i+1\relax}$};
  \foreach \i/\c in {0/cC, 1/cF, 2/cD, 3/cB}
    \node[ndl=\c] (b\i) at ({0.33*(1.5+4*\i)},1.3) {$\mu^{2}_{\the\numexpr\i+1\relax}$};
  \foreach \i/\c in {0/cE, 1/cB, 2/cC, 3/cD, 4/cB, 5/cF, 6/cE, 7/cC}
    \node[ndl=\c] (c\i) at ({0.33*(0.5+2*\i)},0.65) {$\mu^{1}_{\the\numexpr\i+1\relax}$};
  \foreach \i/\c in {0/cC, 1/cD, 2/cB, 3/cE, 4/cF, 5/cC, 6/cD, 7/cB,
                     8/cE, 9/cC, 10/cB, 11/cF, 12/cD, 13/cE, 14/cC, 15/cB}
    \node[tk=\c,minimum size=2.4mm] (d\i) at ({0.33*\i},0) {};
  \foreach \i in {0,1} \draw[ed] (r) -- (a\i);
  \foreach \i/\j in {0/0,0/1,1/2,1/3} \draw[ed] (a\i) -- (b\j);
  \foreach \i/\j in {0/0,0/1,1/2,1/3,2/4,2/5,3/6,3/7} \draw[ed] (b\i) -- (c\j);
  \foreach \i/\j/\k in {0/0/1,1/2/3,2/4/5,3/6/7,4/8/9,5/10/11,6/12/13,7/14/15}{
    \draw[ed] (c\i) -- (d\j);  \draw[ed] (c\i) -- (d\k);}
  \node[lv] at ({0.33*7.5},-0.36) {observed sequence, $16$ tokens};
  \end{scope}

  \newcommand{\minitree}[2]{%
    \foreach \i in {0,...,3}{
      \draw[#2] ({#1*(2*\i+0.5)},0.40) -- ({#1*2*\i},0);
      \draw[#2] ({#1*(2*\i+0.5)},0.40) -- ({#1*(2*\i+1)},0);}
    \foreach \i in {0,1}{
      \draw[#2] ({#1*(4*\i+1.5)},0.80) -- ({#1*(4*\i+0.5)},0.40);
      \draw[#2] ({#1*(4*\i+1.5)},0.80) -- ({#1*(4*\i+2.5)},0.40);}
    \draw[#2] ({#1*3.5},1.20) -- ({#1*1.5},0.80);
    \draw[#2] ({#1*3.5},1.20) -- ({#1*5.5},0.80);
    \foreach \i in {0,...,3} \node[mn] at ({#1*(2*\i+0.5)},0.40) {};
    \foreach \i in {0,1}     \node[mn] at ({#1*(4*\i+1.5)},0.80) {};
    \node[mn] at ({#1*3.5},1.20) {};}
  \newcommand{\rednode}[3]{%
    \pgfmathsetmacro{\px}{#1*(2^#2*#3+2^(#2-1)-0.5)}
    \pgfmathsetmacro{\py}{0.4*#2}
    \pgfmathsetmacro{\cxa}{#1*(2^(#2-1)*2*#3+2^(#2-2)-0.5)}
    \pgfmathsetmacro{\cxb}{#1*(2^(#2-1)*(2*#3+1)+2^(#2-2)-0.5)}
    \pgfmathsetmacro{\cy}{0.4*(#2-1)}
    \draw[emd,cB,line width=0.5pt,shorten <=0.7mm] (\px,\py) -- (\cxa,\cy);
    \draw[emd,cB,line width=0.5pt,shorten <=0.7mm] (\px,\py) -- (\cxb,\cy);
    \node[mn,fill=cB] at (\px,\py) {};}

  \begin{scope}[xshift=1.315cm,yshift=-2.95cm]
  \node[hd] at (0.325,2.0) {(d) The generation probe};
  \minitree{0.20}{emu}
  \draw[emu,cB,line width=0.5pt,shorten >=0.7mm] (0.30,0.80) -- (0.10,0.40);
  \draw[emu,cB,line width=0.5pt,shorten >=0.7mm] (0.30,0.80) -- (0.50,0.40);
  \draw[emu,cB,line width=0.5pt] (0.90,0.40) -- (0.80,0);
  \draw[emu,cB,line width=0.5pt] (0.90,0.40) -- (1.00,0);
  \foreach \i/\c in {0/cB,1/cC,2/cD,3/cE,4/cC,5/cF,6/cB,7/cD}
    \node[tk=\c] at ({0.20*\i},0) {};
  \node[mn,fill=cB] at (0.30,0.80) {};
  \node[mn,fill=cB] at (0.90,0.40) {};
  \node[font=\footnotesize] at (0.70,1.60) {parsed with $\Gpre$};
  \draw[cB,line width=0.5pt] (0.27,0.87) -- (0.27,1.07);
  \draw[cB,line width=0.5pt] (0.85,0.46) -- (0.29,1.05);
  \node[font=\footnotesize,anchor=east] at (0.25,1.20) {deleted rule};
  \node[fbox,inner xsep=3pt] (mdl) at (0.70,-0.80) {fine-tuned model};
  \draw[->,black!55] (mdl.north) -- (0.70,-0.12)
    node[midway,left=1pt,font=\footnotesize,text=black] {generate};
  \end{scope}

  \begin{scope}[xshift=3.70cm,yshift=-2.95cm]
  \node[hd] at (1.70,2.0) {(e) The representation probe};
  \node[font=\footnotesize] at (0.70,1.50) {$x$};
  \node[font=\footnotesize] at (2.70,1.50) {$R\,x$};
  \draw[->,black!55] (1.50,0.70) -- (1.85,0.70) node[midway,above=-1pt,font=\footnotesize,text=black] {$R$};
  \begin{scope}
    \minitree{0.20}{emd}
    \foreach \i/\c in {0/cC,1/cD,2/cB,3/cE,4/cF,5/cC,6/cD,7/cB}
      \node[tk=\c] at ({0.20*\i},0) {};
    \node[circle,draw=black,line width=0.5pt,minimum size=2.4mm,inner sep=0pt] at (1.20,0) {};
  \end{scope}
  \begin{scope}[shift={(2.00,0)}]
    \minitree{0.20}{emd}
    \rednode{0.20}{1}{2}
    \foreach \i/\c in {0/cC,1/cD,2/cB,3/cE,4/cD,5/cB,6/cD,7/cB}
      \node[tk=\c] at ({0.20*\i},0) {};
    \node[circle,draw=black,line width=0.5pt,minimum size=2.4mm,inner sep=0pt] at (1.20,0) {};
  \end{scope}
  \draw[<->,black!55] (1.20,-0.17) -- (1.20,-0.45) -- (3.20,-0.45) -- (3.20,-0.17);
  \node[font=\footnotesize,fill=white,inner sep=1.5pt] at (2.20,-0.45) {$q^{R}$};
  \end{scope}

  \begin{scope}[xshift=7.775cm,yshift=-2.95cm]
  \node[hd] at (2.90,2.0) {(f) The three test families};
  \begin{scope}
    \minitree{0.20}{emd}
    \foreach \i/\c in {0/cC,1/cD,2/cA,3/cE,4/cF,5/cC,6/cD,7/cA}
      \node[tk=\c] at ({0.20*\i},0) {};
    \node[circle,draw=black,line width=0.5pt,minimum size=2.4mm,inner sep=0pt] at (1.40,0) {};
    \node[font=\footnotesize] at (0.70,-0.36) {kept};
  \end{scope}
  \begin{scope}[shift={(2.20,0)}]
    \minitree{0.20}{emd}
    \foreach \l/\i in {3/0, 2/0, 2/1, 1/0, 1/1, 1/2, 1/3}
      \rednode{0.20}{\l}{\i}
    \foreach \i/\c in {0/cE,1/cA,2/cD,3/cC,4/cA,5/cF,6/cE,7/cD}
      \node[tk=\c] at ({0.20*\i},0) {};
    \node[circle,draw=black,line width=0.5pt,minimum size=2.4mm,inner sep=0pt] at (1.40,0) {};
    \node[font=\footnotesize] at (0.70,-0.36) {deleted};
  \end{scope}
  \begin{scope}[shift={(4.40,0)}]
    \minitree{0.20}{emd}
    \foreach \l/\i in {2/0, 1/0, 1/1, 1/2}
      \rednode{0.20}{\l}{\i}
    \foreach \i/\c in {0/cD,1/cC,2/cF,3/cA,4/cE,5/cD,6/cA,7/cC}
      \node[tk=\c] at ({0.20*\i},0) {};
    \node[circle,draw=black,line width=0.5pt,minimum size=2.4mm,inner sep=0pt] at (1.40,0) {};
    \node[font=\footnotesize,align=center] at (0.70,-0.50) {deleted\\in context};
  \end{scope}
  \end{scope}
\end{tikzpicture}
\caption{The Random Hierarchy Model and our setup.
\textbf{(a)} The rule books, one per level, prescribe how each symbol can be expanded in $m$ lower-level synonymic representations. They define the pretraining grammar $\Gpre$. \textbf{(b)} The restriction that defines the grammar used in fine-tuning, $\Gft$: for every symbol on every level, only $m_{\mathrm{keep}}$ rules stay valid while the others get deleted. \textbf{(c)} An example of a sequence with its latent tree structure generated by an RHM with depth $L=4$ and branching factor $s=2$.
\textbf{(d)} to \textbf{(f)} The schematic diagram of the generation probe, the representation probe, and how the sequences in the three test families are generated to measure prediction loss. The gray nodes use the kept rules to generate their children, while the red nodes use the deleted rules.}
\label{fig:rhm-schematic}
\end{figure}
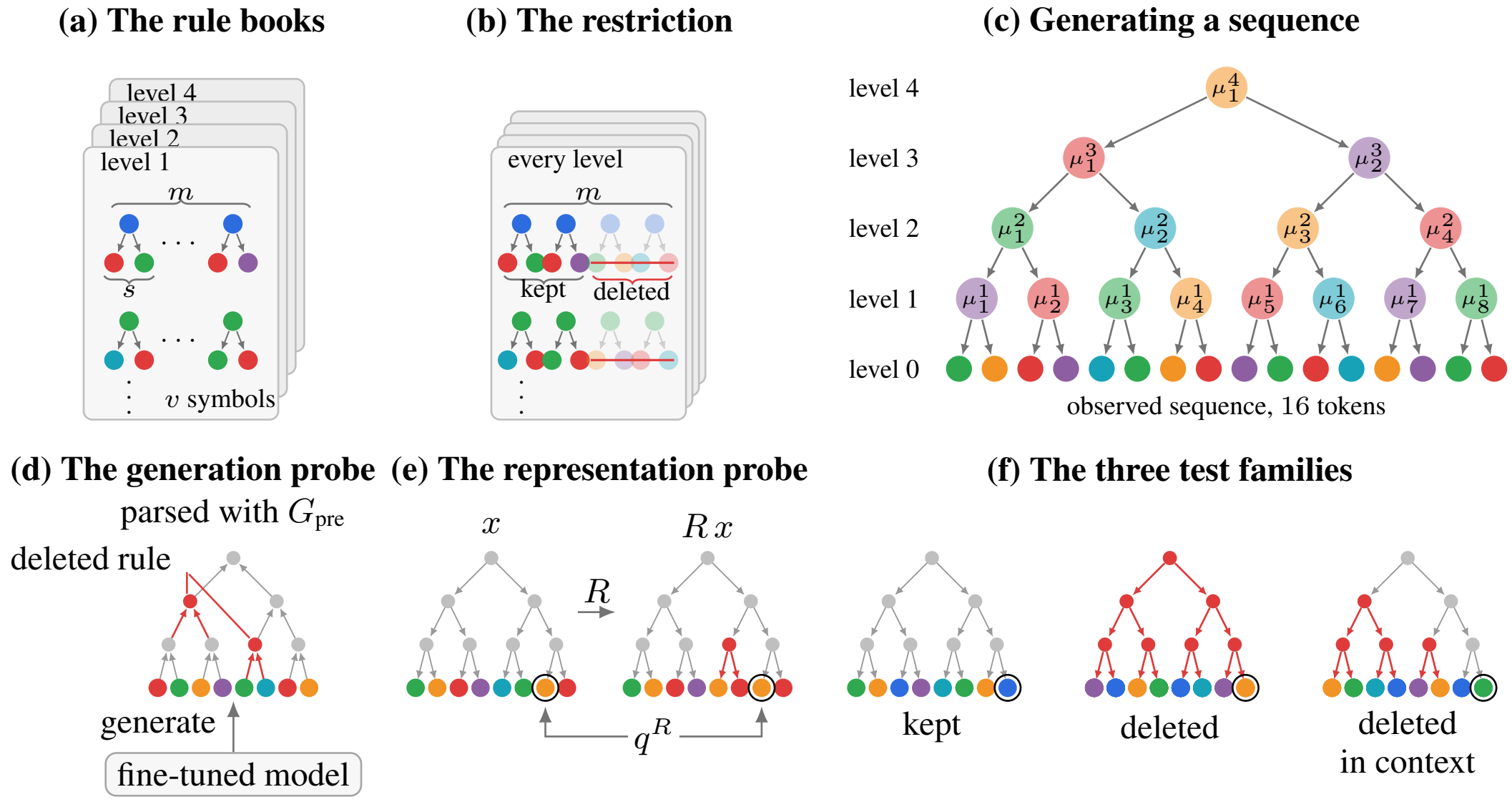

\paragraph{Data model.}
The Random Hierarchy Model (RHM) is a probabilistic context-free grammar that generates sequences through a hierarchy of production rules \citep{cagnetta2024:rhm}. A sequence is obtained by starting from a root node and expanding a regular tree from it, with constant branching factor $s$. The generation proceeds for $L$ levels, until obtaining $s^L \equiv D$ leaf nodes which correspond to the observed tokens. We call this observable level $\ell=0$, while the levels $\ell=1,\dots, L$ are unobserved latent levels (Figure~\ref{fig:rhm-schematic}(a) and (c)). Each node carries one of $v$ symbols. 
One instance of the RHM corresponds to drawing at each level $vm$ distinct ordered $s$-tuples uniformly without replacement from the $v^s$ possible tuples and assigning $m$ to each parent symbol. These alternative expansions are called synonyms. The resulting rule book is shared by all nodes at the same level. No $s$-tuple is assigned to two different parent symbols, so each valid sequence has a unique latent tree. Once the rule books are drawn, they define a grammar $G$ and remain fixed. The grammar generates data by drawing one root symbol uniformly and iteratively sampling one of the synonymic lower-level representations with equal probability. The number of grammatical sequences is $v m^{\frac{D-1}{s-1}}$, therefore exponential in the sequence length $D$.

We write $\mu_i^\ell$ for the $i$th node from the left at level $\ell$. Negative indices count from the right at the same level: $\mu_{-1}^\ell$ is the rightmost node and $\mu_{-2}^\ell$ the second from the right. For example, $\mu_{-2}^2=\mu_3^2$ in the $L=4$, $s=2$ tree. 
The grammar generates the training sequences. The leaf tokens are one-hot encoded in the possible $v$ symbols and used to train a transformer.

\paragraph{Pretraining.}
For a fixed grammar $\Gpre$, we train decoder-only transformers \citep{nanogpt2022} by next-token prediction. Each batch contains $128$ freshly sampled sequences, corresponding therefore to an online learning regime. Training minimizes the mean cross-entropy over the sequence positions with a successor, using Adam \citep{kingma2015adam} at a constant learning rate of $10^{-4}$. 
Architecture, training durations and other settings are given in Appendix~\ref{sec:app-details}.

In our experiments, we use $s=2$ and vary combinations of other values like $L=4$, $v=24$, $m\in\{6,8,10\}$ and $L=3$, $v=16$, $m=8$.

\paragraph{Restricted fine-tuning.}
For the fine-tuning data, we construct $\Gft$ by retaining $m_{\mathrm{keep}} < m$ synonymic representations of every symbol in every rule book and deleting the other rules from $\Gpre$ (Figure~\ref{fig:rhm-schematic}(b)). The retained rule subset is fixed, and the resulting sequences are generated by sampling from the retained synonyms uniformly. Thus, the fine-tuning sequences from $\Gft$ form a narrower distribution of $\Gpre$, while each node still takes one of the possible $v$ values (although its marginal distribution is modified by $\Gft$). Training uses no explicit penalty on deleted rules and no replay of $\Gpre$ data. Starting from the final pretrained weights, we continue next-token prediction on freshly sampled $\Gft$ sequences, with the same optimizer settings and reset Adam moments. The fine-tuning learning rate is $\eta_0=10^{-4}$ unless stated otherwise. 

\subsection{Metrics for forgetting}
\label{sec:metric}

\paragraph{Prediction loss.}
We measure the last-token test cross-entropy $\Llast$, conditioning on all preceding tokens. For pretraining, we compare it with the Bayes-optimal losses $\mathcal{L}_k$, $k=0,\cdots,L$, obtained when only the previous $s^k-1$ context tokens are available. These references are computed by belief propagation on $\Gpre$ \citep{cagnetta2025:ltp2, sclocchi2025phase}; $\mathcal{L}_0$ uses no context and $\mathcal{L}_L$ uses the full sequence context.

For fine-tuning, we evaluate $\Llast$ on three fixed test families, all valid under $\Gpre$ (Figure~\ref{fig:rhm-schematic}(f)). In the \emph{kept-grammar} family, every internal node uses a kept rule. In the \emph{deleted-grammar} family, every internal node uses a deleted rule. In the \emph{deleted-grammar-in-context} family, the $L$ ancestors of the last token use kept rules and all other internal nodes use deleted rules. The last family tests whether the model can interpret deleted rules in the context to predict a kept continuation.

\paragraph{Generation probe.}
We adapt the rule-based evaluation of generated sequences in \citet{favero2025diffusion}. At each checkpoint, we sample the first token from the $\Gft$ marginal and generate the remaining tokens autoregressively at temperature $1$ (\Cref{fig:rhm-schematic}(d)). We parse the sequences with $\Gpre$ and measure the fraction of all nodes at each level that use a deleted rule. 
For a given $m_{\mathrm{keep}}$, the reference deleted-rule rate under $\Gpre$ is $1-\frac{m_{\mathrm{keep}}}{m}$.

\paragraph{Representation probes.}

Following \citet{cagnetta2025:ltp2},
we measure the similarity of the transformer hidden representations from input $x$ and a transformed input $R x$. 

\begin{itemize}
    \item First, we sample $x$ from unmodified $\Gft$.
    \item Then, transformation $R$ replaces the rule from $\Gft$ at the chosen latent node ($\mu_{-2}^ \ell,~\ell=1,2,3$) with a \textit{deleted synonym} while keeping all other nodes to generate the same rules and symbols as $x$ (\Cref{fig:rhm-schematic}(e)). 
    \item For $x$ and $Rx$, we take $\hat h$, a coordinate-wise standardized $d$-dimensional hidden representation after the final normalization layer at the second-to-last token position.
    \item We measure the representation similarity as
    \begin{equation}
    \label{eq:forget-q}
    q^{R}=\frac{1}{d}\,\mathbb{E}_{x,R}
    \left[\hat h(x)\cdot\hat h(Rx)\right],
    \end{equation}
    at each checkpoint throughout fine-tuning.
\end{itemize}

Thus, the metric measures the difference in hidden representations between kept rules and deleted rules when they are supposed to predict the same next token. In fact, for a transformer trained on $\Gpre$, $q^R\simeq 1$ since latent synonyms have the same hidden representations. A decrease in $q^R$ indicates that a deleted rule acquired a different hidden representation with respect to its original group of synonyms.

To set the baseline of $q^R$, we evaluate the same overlap after a symbol replacement: the transformation replaces the symbol at the chosen node by a uniformly drawn different symbol and regenerates its subtree with kept rules. We call this value the \textit{symbol reference}. It is the overlap between inputs whose latents differ, so a decrease of $q^R$ toward it indicates that the representation separates a deleted rule from its kept synonyms as it separates different symbols.

In \Cref{sec:app-probe}, we also introduce \textit{linear probes} as an equivalent measure for the representations of synonymous rules. They bear the same information as $q^R$.

\section{Results}
\subsection{Hierarchical learning in next-token prediction pretraining}
\label{sec:forget-pretrain}

\citet{cagnetta2024:ltp1,cagnetta2025:ltp2} studied pretraining on the RHM for the last-token prediction task, and \cite{favero2025diffusion} studied it for discrete diffusion models. They showed that deep generative models learn to generate sequences consistent with the $\ell$-level grammar rules when the amount of training data reaches the level $P_\ell\sim v m^{\ell+1}$ (or $P_{\ell}\sim v m^{2\ell-1}$ depending on specific architecture and task). Specifically, the value of $P_\ell$ is obtained from a signal-to-noise argument, where the signal strength is given by the correlation strength between $(\ell-1)$-level latent tuples and visible tokens, while the sampling noise strength is given by the law of large numbers.
Since we consider online learning, the number of pretraining time steps $\tau^{(pre)}$ is proportional to seen training samples, and the corresponding training timescales $\tau^{(pre)}_\ell$ to learn level $\ell$ rules follow $\tau^{(pre)}_\ell \propto P_\ell$. 

In Figure~\ref{fig:app-pretrain} of Appendix~\ref{sec:app-pretrain}, we confirm that the next-token prediction task follows the stage-wise learning shown in \cite{cagnetta2025:ltp2, favero2025diffusion}. The transformer's last-token loss decreases through the plateaus $\mathcal{L}_0,\dots,\mathcal{L}_4$, with transitions happening at learning times $\tau_1^{(pre)}, \dots, \tau_L^{(pre)}$ whose scaling is consistent with
\begin{equation}
    \tau_\ell^{(pre)}\propto m^{\ell+1}.
    \label{eq:tpre}
\end{equation}

\subsection{Fine-tuning: two timescales of forgetting}
\label{sec:forget-timescales}

\begin{figure}[t]
    \centering
    \includegraphics{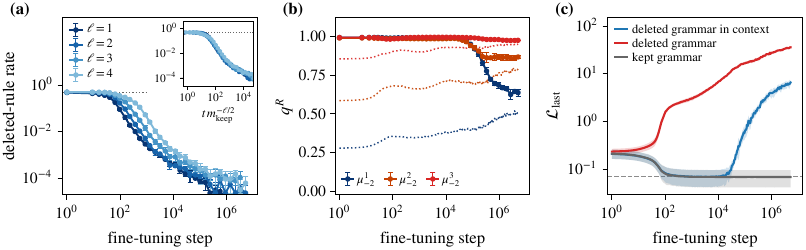}
    \caption{The two timescales of forgetting in the restriction at
    $m = 6$ ($m_{\mathrm{keep}} = 3$); mean $\pm$ std over 4 seeds. We reproduce the same results for $m=8$ and $m=10$~(\Cref{fig:app-timescales}).
    \textbf{(a)} The deleted-rule rate at levels $\ell = 1$ to $4$, against the fine-tuning step $t$, falls from the reference value $1/2$ (dotted), with the curves ordered from the bottom level to the root. Inset: the curves of the four levels, without error bars, collapse against the rescaled step $t\,m_{\mathrm{keep}}^{-\ell/2}$ of \Cref{eq:tft} (full size in Figure~\ref{fig:app-level-collapse}).
    \textbf{(b)} The representation probe $q^{R}$ at $\mu_{-2}^{1}$, $\mu_{-2}^{2}$ and $\mu_{-2}^{3}$ stays close to $1$ long after generation is suppressed, and starts decreasing toward each node's symbol reference (dotted) only orders of magnitude of steps later.
    \textbf{(c)} $\Llast$ on the three test families; dashed line: the Bayes-optimal loss of $\Gft$, seed mean. The deleted-grammar loss rises during the suppression, while the deleted-grammar-in-context loss stays close to the Bayes-optimal loss until $q^R$ decreases.}
    \label{fig:forget-timescales}
\end{figure}

Restriction fine-tuning rapidly suppresses the generation of deleted rules, while their representations and their predictive use in context persist for much longer.

\paragraph{Rapid suppression of generation.}
Figure~\ref{fig:forget-timescales}(a) shows that the deleted-rule rate falls from near $1/2=1-m_{\mathrm{keep}}/m$ to near zero at every level and remains low throughout fine-tuning. Suppression proceeds from the bottom rule book ($\ell=1$) to the root ($\ell=4$), with the same ordering across seeds. Unparseable sequences due to grammatical errors occur during the initial fine-tuning but become rare later, so the decrease in deleted-rule use is accompanied by grammatically valid generations. 

In the inset of Figure~\ref{fig:forget-timescales}(a) and in Figure~\ref{fig:forbidden-L3}, we plot the generation curves against $t m_{\mathrm{keep}}^{-\ell/2}$, where $t$ counts fine-tuning steps. Figure~\ref{fig:app-level-collapse} shows the rescaled curves at full size for $m=6$, $8$ and $10$.
The collapse shows that the fine-tuning timescale for the generation drop is consistent with 
\begin{equation}
    t^{(ft)}_{\ell} \sim m_{\mathrm{keep}}^{\ell/2},
    \label{eq:tft}
\end{equation}
which is inversely proportional to the magnitude of the correlations between a token and its level-$\ell$ latent ancestor for an RHM with $m=m_{\mathrm{keep}}$ \citep{cagnetta2024:ltp1}. In \Cref{sec:theory} and \Cref{app:theory}, we show that these correlations set the magnitude of the first fine-tuning gradient update, therefore suggesting that this signal strength determines the timescale for generation change during fine-tuning.

\paragraph{Delayed change in representations.}
The representation probe $q^R$ remains close to $1$ after generation has been suppressed at every level (Figure~\ref{fig:forget-timescales}(b)). During this period, replacing a kept rule in context with a deleted synonym has little effect on the representation used for prediction. The model therefore retains approximate invariance to deleted expressions of the same latent despite rarely generating them.

Only after orders of magnitude more fine-tuning does $q^R$ decrease toward the symbol reference. 
The probe $q^R$ at $\ell=1,2,3$ starts to decrease at similar times, in contrast to the clear sequential ordering of learning each hierarchical latent level during pretraining (Figure~\ref{fig:app-pretrain}(b)).

\paragraph{Predictive use of deleted rules in context.}
In Figure~\ref{fig:forget-timescales}(c), the three test families distinguish suppression of deleted continuations from the loss of the ability to interpret deleted rules in context. The kept-grammar loss decreases smoothly toward its Bayes-optimal value.
The deleted-grammar loss rises during the initial suppression of generation and continues to increase.

The deleted-grammar-in-context sequences require the model to interpret deleted rules in the context while predicting a continuation in kept rules. Their loss initially decreases together with the kept-grammar loss and remains close to the $\Gft$ Bayes-optimal value for a much longer period after the rapid increase of the deleted-grammar loss. This indicates that the deleted rules remain useful for prediction even after the behavioral suppression. At a much longer timescale, instead, we see the loss of deleted-grammar-in-context increase, indicating that the predictive use of deleted rules deteriorates. We observe that the timescale where this increase begins coincides with the decrease of the representation probe $q^R$ (Figure~\ref{fig:forget-timescales}(b)). 

\subsection{Hierarchical correlations and the fast behavioral suppression}
\label{sec:theory}

In this section, we propose an origin for the observed fine-tuning timescales of suppressed generation (\Cref{sec:forget-timescales}), summarized in \Cref{eq:tft}. In line with existing theories of RHM learnability, we relate the strength of the learning signal to the magnitude of token--latent correlations, observing that they are simply related to \Cref{eq:tft}.

Building on the observation in \Cref{sec:forget-timescales} that the internal representations stay unchanged in the early fine-tuning dynamics, in \Cref{app:toy-model} we consider a linear model on fixed features that learned to encode the latent data structure during pretraining. In this simplified setting, the first fine-tuning gradient step, which we call \textit{fine-tuning (f.t.) signal}, has size
\[
\text{f.t. signal} = \Delta p(y\mid x)
=
p_{\mathrm{ft}}(y\mid x)
-
p_{\mathrm{pre}}(y\mid x),
\]
where $p_{\mathrm{pre}}(y\mid x)$ and $p_{\mathrm{ft}}(y\mid x)$ are respectively the pretraining and fine-tuning distributions of next-token $y$ conditioned on context $x$.

\paragraph{Hierarchical scaling of the fine-tuning signal.}
In the RHM, because of its tree structure, the conditional probabilities $p_\nu(y\mid x)$, where
$\nu\in\{\mathrm{pre},\mathrm{ft}\}$, are mediated by the hidden latents $h(x)$ identified by the context $x$: $p_\nu(y\mid x) = p_\nu(y\mid h(x))$.
Therefore, changing the generation probabilities corresponding to a latent rule at level $\ell$ corresponds to changing the output in the direction of $p_{\mathrm{ft}}(y\mid h^{(\ell)}(x))-p_{\mathrm{pre}}(y\mid h^{(\ell)}(x))$, where $h^{(\ell)}(x)$ is the lowest common ancestor between the context $x$ and the next-token $y$.

In \Cref{app:conditional-change}, we compute the magnitude of $p_{\mathrm{ft}}(y\mid h^{(\ell)}(x))-p_{\mathrm{pre}}(y\mid h^{(\ell)}(x))$ with respect to the RHM parameters and $m_{\mathrm{keep}}$ and hierarchical level $\ell$. We obtain $\Delta p(y\mid h^{(\ell)}(x))  \sim\frac{\sqrt{1-(m_{\mathrm{keep}}/m)^\ell}} {\sqrt v\,m_{\mathrm{keep}}^{\ell/2}}$, that for $m_{\mathrm{keep}}\ll m$ has the scaling 
\begin{equation}
\Delta p(y\mid h^{(\ell)}(x)) 
\sim
1/m_{\mathrm{keep}}^{\ell/2}.
\label{eq:restriction-signal}
\end{equation}
Assuming that the fine-tuning timescales $t_{\ell}^{(ft)}$ follow
\begin{equation}
    t_{\ell}^{(ft)}\sim \frac{1}{\text{f.t. signal}},
    \label{eq:assumption}
\end{equation} 
one obtains $t^{(ft)}_{\ell} \sim m_{\mathrm{keep}}^{\ell/2}$ consistently with the observations of \cref{eq:tft}.

\paragraph{Limitations and tests of the assumption of \cref{eq:assumption}.}
\Cref{eq:assumption} assumes that \textit{(i)} the model initially behaves as a linear readout from existing features, and that \textit{(ii)}
its first step persists across the first part of the fine-tuning dynamics, until the output reaches a margin where the loss and its gradients eventually start changing. This assumption is justified only for the case $\ell=1$, i.e., changing the token probability w.r.t. its nearest neighbor token, because \textit{(i)} this conditional probability can be represented by a linear readout and \textit{(ii)} before the linear model learns this change, the average loss and its gradients are effectively constant. 
However, for higher hierarchical levels, both conditions \textit{(i)} and \textit{(ii)} are not valid: \textit{(i)} output probabilities depend non-linearly on combinations of latent variables and \textit{(ii)} the loss has started evolving after changing $p(y\vert h^{\ell=1})$.

In \Cref{app:linearReadout}, we report the behavior of fine-tuning dynamics for the linearized model  
\(f_{\mathrm{lin}}(x;\Delta\theta)=f_{\theta_0}(x)+J_{\theta_0}(x)\Delta\theta\), where $J_{\theta_0}(x)$ is the Jacobian of the transformer after pretraining. In \Cref{fig:liner_nonliner}, we observe that, consistently with this discussion, the behavior timescale for $\ell=1$ is the same as that of the full non-linear model, and eventually starts changing after the drop in $p(y\vert h^{\ell=1})$ for the deleted rules.
Therefore, understanding why \cref{eq:assumption} is consistent with the observations (\Cref{eq:tft}) beyond the initial fine-tuning dynamics requires a better understanding of the full non-linear dynamics, which remains an open problem.

\paragraph{Difference with pretraining complexity argument.}
Since we are considering online training with fixed batch size, the number of training steps is proportional to the number of observed samples. Therefore, the training steps necessary for learning are (at least) proportional to the sample complexity.
Theoretical arguments for RHM pretraining sample complexity \citep{cagnetta2024:ltp1,favero2025diffusion} compare the learning signal magnitude, i.e., the latent--token correlations, with their sampling noise in their empirical estimates, obtaining \Cref{eq:tpre}, which scales differently from \Cref{eq:tft}.
We think the main difference between pretraining and fine-tuning dynamics is that fine-tuning reuses the latent representations acquired during pretraining, and therefore adaptation does not need to first reconstruct the latent hierarchy.
A more precise understanding of the observed different timescale behaviors requires a deeper analysis of the learning dynamics of both pretraining and fine-tuning.

\subsection{Noisy representational drift and the slow timescale}
\label{sec:forget-weights}

\begin{figure}[t]
    \centering
    \includegraphics{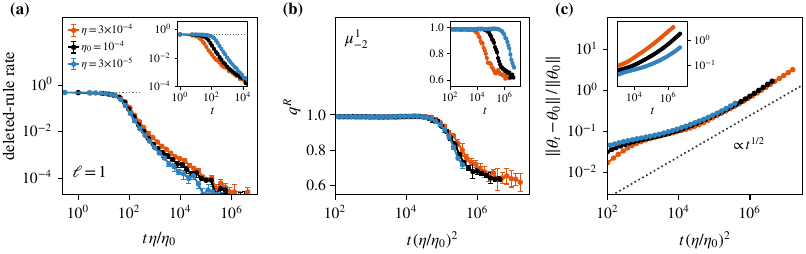}
    \caption{Restriction fine-tuning at $m=6$, $m_{\mathrm{keep}}=3$ with learning rates
    $\eta=3\times10^{-4}$, $\eta_0=10^{-4}$ (the learning rate of all other experiments) and
    $3\times10^{-5}$; mean $\pm$ std over $3$ seeds. The insets show the curves averaged over seeds vs the fine-tuning step $t$.
    \textbf{(a)} Deleted-rule rate at level $\ell=1$
    : the curves decrease from the starting value $1/2$ (dotted line) at different times depending on $\eta$ (inset), and they collapse when plotted vs $t\,\eta/\eta_0$. This is consistent with a suppression of generation driven by the mean weight updates.
    \textbf{(b)} Probe $q^R$ at $\mu_{-2}^1$: also the time-scale of representational forgetting depends on the learning rate $\eta$ (inset), but
    curves collapse when plotted vs $t\,(\eta/\eta_0)^2$, differently from panel (a). This is consistent with representational forgetting being driven by accumulated noise in the weight updates.
    \textbf{(c)} Distance from the pretrained parameters relative to their norm, $\|\theta_t-\theta_0\|/\|\theta_0\|$: 
    consistently with panel (b), the late time displacement depends on $\eta$ (inset) and the curves collapse when plotted vs $t\,(\eta/\eta_0)^2$, converging to an asymptotic displacement $\propto t^{1/2}$ (dashed line).
    }
    \label{fig:forget-lr}
\end{figure}

Section~\ref{sec:theory} relates the timescale of behavioral suppression to the fine-tuning gradient.
As discussed in Section~\ref{sec:forget-timescales}, the fast behavioral suppression is followed by a later regime of representational forgetting.
Here we show that this slow change in representations is consistent with representational drift caused by noise in the parameter updates.

Our hypothesis is that the average gradient of the fine-tuning optimization has no component that impairs or conserves the deleted-rule representations. Instead, their degradation at large times is due to the accumulation of random displacements in the weight updates caused by the variance in the batch updates.

\paragraph{Test 1: Varying learning rate $\eta$}
The hypothesis above predicts a different dependence on the learning rate $\eta$ for the fast behavioral timescale and the slow representational one. Adam normalizes each update by a running estimate of its gradient magnitude \citep{kingma2015adam}, so that the magnitude of each update is approximately of order $\eta$, whatever the magnitude of the gradient. If successive updates of each parameter have the same sign, as under a persistent gradient, its displacement after $t$ steps is proportional to $\eta\,t$. If their signs fluctuate, as under noise, the displacement grows as $\eta\,t^{1/2}$. Therefore, a given displacement takes a number of steps proportional to $\eta^{-1}$ when caused by a persistent gradient, and to $\eta^{-2}$ when caused by accumulated noise.

We test this prediction by varying the learning rate (Figure~\ref{fig:forget-lr}). The curves of the deleted-rule rate at the three learning rates coincide when plotted against $t\,\eta/\eta_0$ (Figure~\ref{fig:forget-lr}(a); all levels in Figure~\ref{fig:app-lr-gen}). The curves of $q^R$ coincide when plotted against $t\,(\eta/\eta_0)^2$ (Figure~\ref{fig:forget-lr}(b); both scalings in Figure~\ref{fig:app-lr-slow}). A linear probe that reads the symbol of the probed node from the hidden representation (Appendix~\ref{sec:app-probe}) gives the same result (Figure~\ref{fig:app-lr-probe}). 
Figure~\ref{fig:forget-lr}(c) shows the displacement of the parameters relative to their pretrained values $\theta_0$. In the later stage, the increase is approximately proportional to $t^{1/2}$, and the curves of the three learning rates coincide when plotted against $t\,(\eta/\eta_0)^2$ (see also Figure~\ref{fig:app-lr-slow}), consistent with the behavior of $q^R$.

\paragraph{Test 2: Isolating the gradient variance}
Another test for our hypothesis is removing the gradient average after the behavioral forgetting, so that only its variance can drive the late-stage dynamics.
For this test, we use a modified loss that clips low probabilities to zero, which we call \textit{bounded suppression}: a token other than the target is discarded from the cross-entropy once its logit lies more than a fixed margin below the largest logit at its position. 
In this way, tokens that are suppressed beyond a margin do not contribute to the loss and its gradients.

With this loss, in Figure~\ref{fig:app-bounded}(a) to (d) of Appendix~\ref{sec:app-bounded}, we observe that the deleted-rule rate stops decreasing at a small value, consistent with a vanishing average update. 
Instead, in Figure~\ref{fig:app-bounded}(e), $q^R$ follows the same curve as the baseline training, consistent with a dynamics that is driven by the gradient variance. The linear probe gives the same result (Figure~\ref{fig:app-bounded}(f)).
The slow change in representations is therefore not caused by the continued suppression of deleted rules but by the variance of the weight updates.

\paragraph{The role of explicit regularizations} Finally, we also show that two explicit regularizations can alter the slow timescale while keeping the initial suppression of generation unchanged (Appendix~\ref{sec:app-wd}). An $L_2$ regularization toward the pretrained parameters $\theta_0$, applied separately from the Adam update, restricts the late motion of the parameters, and can even retain the representation of the deleted rules (\Cref{fig:app-wd}(d) to (f)). 
Instead, weight decay \citep{loshchilov2019decoupled} with coefficient $\lambda$ is mathematically equal to the same regularization toward zero, and hence moves the parameters away from $\theta_0$ over a time of order $1/(\eta\lambda)$. When this time is shorter than the timescale of the change without decay, the representations are lost after a time of order $1/(\eta\lambda)$ (\Cref{fig:app-wd}(a) to (c)).

\FloatBarrier
\section{Experiments with natural language data}
\label{sec:realistic}

We primarily use RHM as a controlled synthetic setup to discover and understand the dissociation between the timescales of behavioral and latent forgetting. To connect these findings to more naturalistic data, we devise experiments with a decoder-only transformer trained with a bilingual natural language corpus. 

\suppressfloats[t]
\begin{figure}[!htbp]
    \centering
    \includegraphics[width=\linewidth]{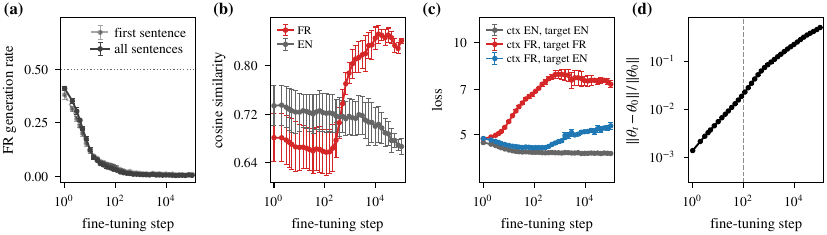}
    \caption{Bilingual pretraining followed by EN-only fine-tuning.
    \textbf{(a)} FR generation rate drops rapidly before the $10^2$ fine-tuning steps, both for the first sentence and all sentences in the context window. \textbf{(b)} Cosine similarity between
    representations of coherent and token-shuffled sentences in FR and EN; higher similarity indicates loss of latent linguistic capability. Rapid rise of FR happens after $10^2$ steps. \textbf{(c)} Reproducing the three test families (\Cref{fig:rhm-schematic}(f)). Prediction losses for EN targets with EN context (gray), FR targets with FR
    context (red), and EN targets with FR context (blue) show a similar trend, characterized by a further drop of the loss for gray (kept rules), a faster rise of the loss in red (deleted rules) and a slower timescale of blue (deleted rules in context).
    \textbf{(d)} Parameter distance relative to the pretrained norm; the vertical line marks $10^2$ steps, approximately coinciding with the transition in the behavioral and representational forgetting.}
    \label{fig:realistic}
\end{figure}

For pretraining, we train an 8-layer decoder-only transformer on a corpus in which each topically coherent passage is written in either English (EN) or French (FR) sentences, chosen at random. This is a naturalistic reflection of RHM where the latent contains the grammatical rules for generating English or French text. We then fine-tune on a monolingual (EN or FR) corpus, mirroring the fine-tuning on restricted generative rules in RHM. Finally, we evaluate the same measures of behavioral and latent forgetting in this model. Details of the pretraining and fine-tuning corpus generation, training setup, and evaluation are in~\Cref{app:realistic-bilingual}. 

We focus on the model fine-tuned on EN-only text here, but the results are consistent with the FR-only fine-tuned model (\Cref{fig:app-realistic-fr}). After fine-tuning on EN-only text, we replicate the measurement setting used for RHM in~\Cref{fig:forget-timescales}. In \Cref{fig:realistic}(a), we find that the generation rate of FR drops rapidly long before $10^2$ steps. To test whether the latent representations of one language in the model are intact, we measure the representational similarity between coherent sentences and incoherent versions with randomly shuffled tokens. High similarity indicates that the model lost capacity to encode coherent meaning of the language, a sign of representational forgetting (\Cref{fig:realistic}(b)), which happens only after $10^2$ steps.  In \Cref{fig:realistic}(c), we replicate the measurement of~\Cref{fig:rhm-schematic}(f) in the bilingual model, again marking the transition in generation suppression and representational forgetting at $10^2$ steps. Finally, we also observe a slight crossover in the increase of the norm of relative weight change (\Cref{fig:realistic}(d)), showing that also in this setting the model weights keep drifting even when the output has adapted to the fine-tuning distribution. A further control experiment on context ablation shows that after the onset of representational forgetting, the language removed during fine-tuning obtains almost zero loss gap between the informative context and uninformative context while the kept language gains a significantly lower loss given the informative context (\Cref{fig:app-realistic-controls-en}(b)). Including the removed language after the fine-tuning also shows that the relearning ability significantly drops after the onset of representational forgetting while beforehand, the model immediately recovers the removed language even after the behavioral forgetting (\Cref{fig:app-realistic-controls-en}(c)). We reproduce a similar observation with the English-Turkish language pair (\Cref{fig:app-realistic-wd-entr-en,fig:app-realistic-wd-entr-tr}). They have substantially different syntax and less shared structured than English-French and this supports that our result does not rely on highly shared latent structure. 

Finally, to show that our observations persist in models at scale, we perform targeted fine-tuning on open-source code generation LLMs, Qwen2.5-Coder~\citep{hui2024qwen2} 0.5B and 1.5B. We fine-tune on specific formatting syntax (indentation, string formatting) among many plausible ones, and show similar timescale separations in limited cases. See~\Cref{app:realistic-coding} for the results and the experimental details. 

Overall, these experiments show that the separation of timescales between behavioral and representational forgetting is not restricted to the synthetic RHM but is a general phenomenon in fine-tuning with natural language.

\section{Discussion}

Our results show that the outcome of fine-tuning on restricted data is governed by two timescales: a fast timescale of behavioral forgetting in outputs, consistent with the latent--token correlations, and a slow timescale of representational forgetting, stemming from the noise accumulated over the fine-tuning updates. 
This separation of timescales provides an explanation for observations in LLMs where capabilities suppressed by fine-tuning or unlearning can still be elicited through alternative prompts or recovered from hidden representations
\citep{kotha2024understanding,zheng2025spurious,lynch2024eight}: such models are in an intermediate regime between the two timescales, in which behavior has adapted while the underlying representations remain largely intact. Beyond this initial regime, our findings highlight how the accumulation of noise in the parameter updates can drive a slow representational drift~\citep{blanc2020implicit, li2021happens}, eventually leading, on much longer timescales, to the loss of representations supporting the suppressed capabilities.

In this work, the considered setting is constrained to a specific form of fine-tuning, that is a narrower distribution of the training data. In~\Cref{sec:app-resample}, we show that fine-tuning with out-of-distribution data generated from new sets of rules shows forgetting-relearning dynamics. We leave a thorough analysis of this case for future work. 

While here we focus exclusively on SFT, Reinforcement Learning (RL) is another important and widely used fine-tuning method in modern training pipelines. Studying how knowledge and skills from pretraining can continue to be learned, composed, and reused throughout RL post-training~\citep{lee2024animals, zhao2025echo} is an exciting future direction.

\section*{Acknowledgments}
We thank Francesco Cagnetta, Matthieu Wyart, Yedi Zhang and Nishil Patel for valuable discussions.
JHL, AS and AMS thank the Sainsbury Wellcome Centre Core Grant from Wellcome (219627/Z/19/Z) and the Gatsby Charitable Foundation (GAT3850). All authors thank the Coefficient Giving for Technical AI safety grant. 

\section*{Ethics Statement}

While our work aims for a fundamental understanding of language models and their fine-tuning process, we do not anticipate any immediate societal impact from this research. 

\section*{Reproducibility Statement}
We report detailed experimental setup to reproduce all the results in the Appendix. The code to reproduce the main experiment on the Random Hierarchy Model (Figure~\ref{fig:forget-timescales}) is available at \url{https://github.com/yilin-luo/two-timescales-of-forgetting}.

\FloatBarrier
\bibliography{iclr2027_conference}
\bibliographystyle{iclr2027_conference}

\FloatBarrier
\appendix
\crefalias{section}{appendix}\crefalias{subsection}{appendix}
\section{Additional results and experimental details on the Random Hierarchy Model}

\subsection{Experimental details}
\label{sec:app-details}

The settings below apply to the restriction experiments with $L=4$,
$s=2$, $v=24$, $m\in\{6,8,10\}$ and $m_{\mathrm{keep}}=m/2$.
We retain the last $m_{\mathrm{keep}}$ entries of each symbol's rule
list. Since the entries are exchangeable under the grammar draw, this
is equivalent to selecting a uniformly random subset.
Table~\ref{tab:app-settings} supplements the training setup in
Section~\ref{sec:forget-setup}.

For the representation probes of Section~\ref{sec:metric}, both inputs
are standardized using the coordinate-wise mean and standard deviation
of the untransformed samples at the same checkpoint. After replacing
a rule, we regenerate its descendants using kept rules; tokens outside
the selected subtree are unchanged. The symbol reference uses the same
original samples as $q^R$. In the generation probe, unparseable
nodes remain in the denominator and count as neither kept nor deleted.

\begin{table}[!htb]
\caption{Network, training and evaluation settings supplementing
Section~\ref{sec:forget-setup} for the $L=4$ restriction experiments.}
\label{tab:app-settings}
\centering
\begin{tabular}{lp{3.35in}}
\toprule
blocks & $4$, with layer normalization before each attention and feed-forward operation \\
hidden width $d$, heads & $512$, $16$ \\
feed-forward layers & hidden width $4d$, Gaussian-error linear unit (GELU) \\
positional embedding & learned, absolute \\
readout & linear, tied to the token embedding, after a final layer normalization \\
biases, dropout & none \\
parameters $n_\theta$ & $12.6\times10^6$ \\
\midrule
Adam coefficients & $\beta_1=0.9$, $\beta_2=0.999$ \\
weight decay & none in baseline runs; decay in Appendix~\ref{sec:app-wd} applies to matrices and embeddings, not layer-normalization gains \\
fine-tuning learning rate & $\eta_0=10^{-4}$; $3\times10^{-5}$ and $3\times10^{-4}$ in Figure~\ref{fig:forget-lr} and Appendix~\ref{sec:app-lr} ($m=6$) \\
gradient clipping & none \\
test set & $2^{15}$ held-out sequences \\
pretraining steps & $10^6$ at $m=6$; $10^7$ at $m=8$ and $10$ \\
fine-tuning steps & up to $10^7$, continuing beyond the initial convergence of the kept-grammar loss \\
fine-tuning checkpoints & nominal $280$-point logarithmic grid, rounded to integer steps with duplicates removed; weights only \\
seeds & $4$ per $m$, each with its own grammar and network initialization; $3$ in the learning-rate comparison \\
\midrule
generation probe & $8192$ sequences per checkpoint \\
representation probe & $4096$ original sequences per node and checkpoint \\
linear probe & $4096$ training and $4096$ test sequences per node and checkpoint \\
test families & $8192$ sequences per family and seed \\
\bottomrule
\end{tabular}
\end{table}

\FloatBarrier
\subsection{Pretraining: network depth and scaling across grammar sizes}
\label{sec:app-pretrain}

We supplement Section~\ref{sec:forget-pretrain} with a network-depth
comparison and a test across grammar sizes. The loss curves show
$\Llast$ against the number of sampled training sequences $P$. For
the plateau references defined in Section~\ref{sec:metric}, belief
propagation marginalizes all leaves outside the specified context,
following Eq.~22 of \citet{cagnetta2025:ltp2}. Each reference is
evaluated using the seed's own grammar and $8192$ held-out sequences.

The figures mark the sample scales of Section~\ref{sec:forget-pretrain}
with their prefactor,
\begin{equation}
\label{eq:app-pretrain-scales}
P_1\propto(1-f)^{-1}v\,m,
\qquad
P_\ell\propto(1-f)^{-1}v\,m^{\ell+1}\quad(\ell\geq2),
\qquad
f=\frac{m}{v^{s-1}}.
\end{equation}
The bottom book is the exception to the form $m^{\ell+1}$. No latent
below it has to be inferred, and learning it only requires observing
its $vm$ admissible $s$-tuples \citep{favero2025diffusion}; $P_1$ is the
first scale of Eq.~12 in \citet{cagnetta2024:ltp1}.

During pretraining, we evaluate the overlap defined in \eqref{eq:forget-q} under
the symbol replacement that defines the symbol reference in
Section~\ref{sec:metric}, sampling $x$ from $\Gpre$ and regenerating
the replaced subtree with $\Gpre$. We denote this probe by $q^S$.
It is close to $1$
when the representation does not depend on the latent at that node and
decreases as this dependence develops. At the end of pretraining at
$m=6$, its values are close to those of the symbol reference at the
start of fine-tuning in Figure~\ref{fig:forget-timescales}(b).

\begin{figure}[!htb]
    \centering
    \includegraphics{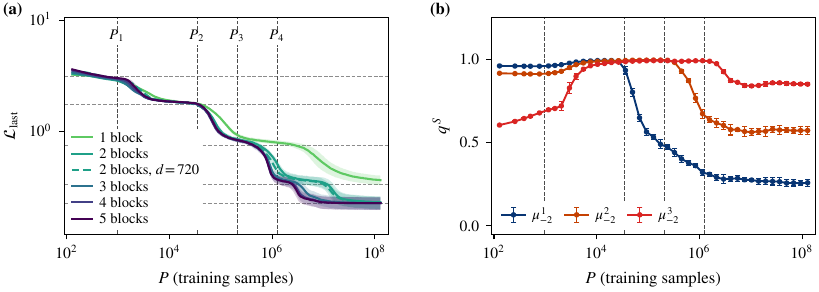}
    \caption{Pretraining at $m=6$.
    \textbf{(a)} $\Llast$ for networks with $1$ to $5$ blocks,
    and a two-block network with hidden width $720$ (dashed,
    labeled $d=720$ in the legend), approximately matching the
    parameter count of the four-block network; mean $\pm$ std over
    $4$ seeds. Horizontal dashed lines: the seed-averaged plateau
    losses $\mathcal L_0,\dots,\mathcal L_4$.
    Vertical dashed lines: the sample scales $P_1,\dots,P_4$ of
    \eqref{eq:app-pretrain-scales}. Their common prefactor is fixed
    by the geometric mean, over the five network depths, of the
    sample count where the mean loss crosses $\mathcal L_1$.
    \textbf{(b)} $q^S$ for the four-block network at
    $\mu_{-2}^1$, $\mu_{-2}^2$ and $\mu_{-2}^3$;
    mean $\pm$ std over $8$ seeds, with the same vertical lines.}
    \label{fig:app-pretrain}
\end{figure}

The depth comparison uses the same grammars and training settings,
with $10^6$ steps for each network. Reducing the number of blocks
delays the later loss decreases, and the one-block network does not
reach the full-context reference within the training duration.
Increasing the width of the two-block network leaves its learning
curve close to that of the narrower two-block network. Thus, the
delay in these runs cannot be attributed solely to their smaller
parameter count.

\begin{figure}[!htb]
    \centering
    \includegraphics{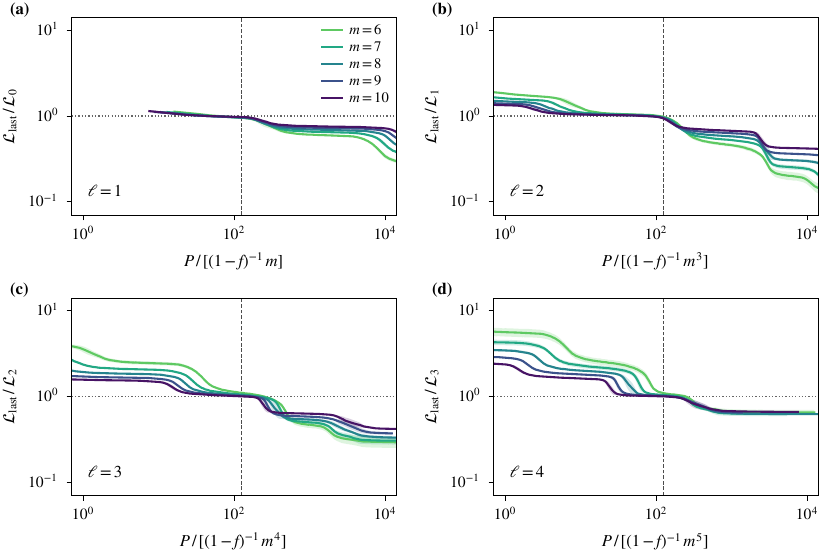}
    \caption{Last-token learning curves of the four-block network
    at $m=6,\dots,10$; mean $\pm$ std over $8$ seeds.
    Panels \textbf{(a)} to \textbf{(d)} correspond to the transitions
    $\ell=1,\dots,4$. Each plots $\Llast/\mathcal L_{\ell-1}$, using the
    seed's own plateau reference, against $P$ divided by the
    $m$-dependent factor of $P_\ell$ in \eqref{eq:app-pretrain-scales}.
    Dotted horizontal line: $1$. Dashed vertical line: the common
    prefactor fixed in Figure~\ref{fig:app-pretrain}, expressed
    on the rescaled axis.}
    \label{fig:app-rescaled}
\end{figure}

The comparison across grammar sizes uses $10^6$ pretraining steps
at $m=6,7,9$ and $10^7$ at $m=8,10$.
Under the rescaling in Figure~\ref{fig:app-rescaled}, the departures
from successive plateaus approximately align across $m$.
The agreement concerns the transition locations; the losses reached
after each transition remain grammar-dependent.

\FloatBarrier
\subsection{Additional results}
\label{sec:app-results}

\begin{figure}[!htb]
    \centering
    \includegraphics{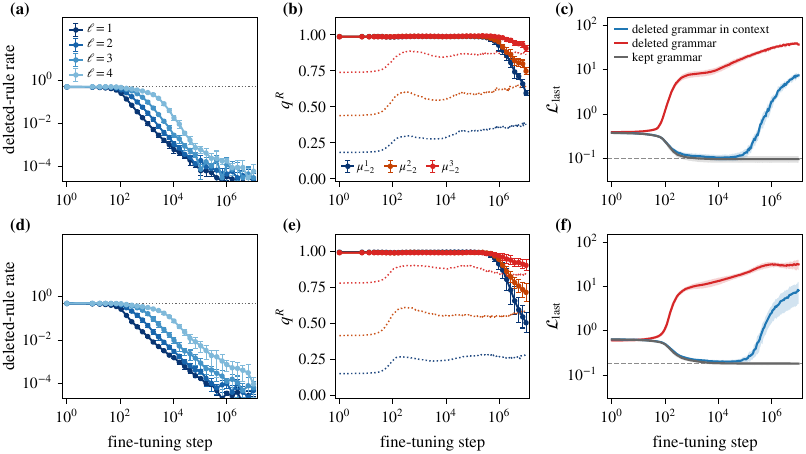}
    \caption{Restriction fine-tuning at $m=8$,
    $m_{\mathrm{keep}}=4$ (top row), and $m=10$,
    $m_{\mathrm{keep}}=5$ (bottom row); mean $\pm$ std over
    $4$ seeds. \textbf{(a), (d)} Deleted-rule generation rates at
    levels $\ell=1,\dots,4$ against fine-tuning step $t$; dotted
    line: $1/2$. The same curves against
    $t\,m_{\mathrm{keep}}^{-\ell/2}$ are shown in
    Figure~\ref{fig:app-level-collapse}(b), (c).
    \textbf{(b), (e)} $q^R$ at $\mu_{-2}^1$, $\mu_{-2}^2$ and
    $\mu_{-2}^3$, with each node's symbol reference (dotted).
    \textbf{(c), (f)} $\Llast$ on the three
    test families; gray dashed line: the seed-averaged Bayes-optimal
    loss of $\Gft$. Colors follow Figure~\ref{fig:forget-timescales}.}
    \label{fig:app-timescales}
\end{figure}

Figure~\ref{fig:app-timescales} extends
Figure~\ref{fig:forget-timescales} to $m=8$ and $m=10$.
Representational change occurs earlier at $m=6$ than at these two
grammar sizes, whose timescales are similar. However, their base
networks were pretrained for longer (Table~\ref{tab:app-settings})
and have larger parameter norms. This comparison therefore does not
isolate the dependence of the slow timescale on $m$.

\begin{figure}[!htb]
    \centering
    \includegraphics{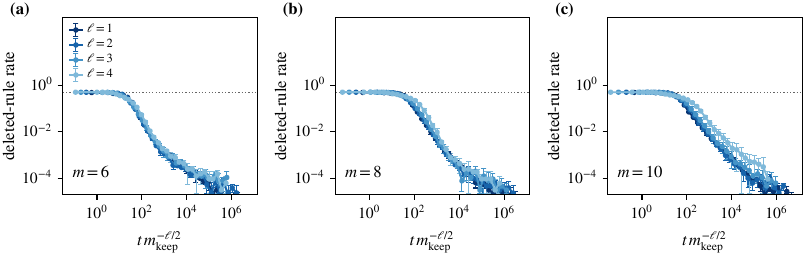}
    \caption{Deleted-rule rate at levels $\ell=1$ to $4$ against the rescaled step
    $t\,m_{\mathrm{keep}}^{-\ell/2}$ in the restriction fine-tuning at \textbf{(a)} $m=6$,
    $m_{\mathrm{keep}}=3$, \textbf{(b)} $m=8$, $m_{\mathrm{keep}}=4$ and \textbf{(c)} $m=10$,
    $m_{\mathrm{keep}}=5$; mean $\pm$ std over $4$ seeds; dotted line: $1/2$. Under this rescaling,
    the curves of the four levels, which start to fall at different steps in
    Figures~\ref{fig:forget-timescales}(a) and~\ref{fig:app-timescales}(a), (d), collapse,
    consistent with \cref{eq:tft}.}
    \label{fig:app-level-collapse}
\end{figure}

\begin{figure}[!htb]
    \centering
    \includegraphics{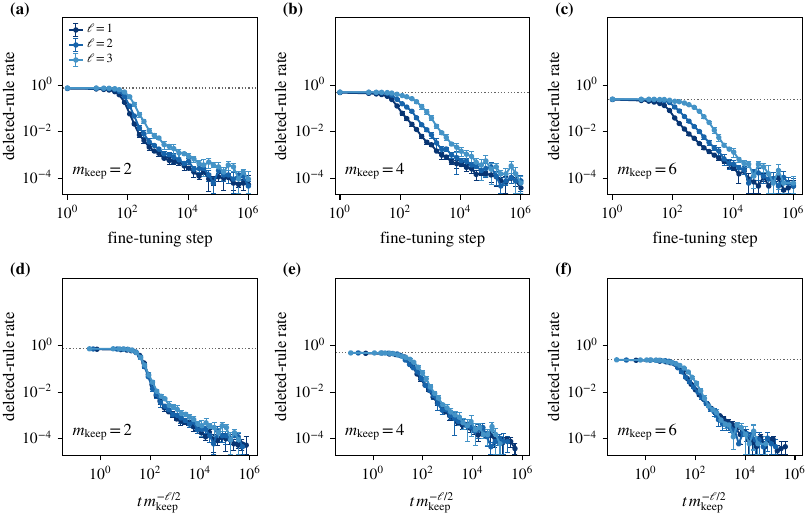}
    \caption{Restriction fine-tuning for RHM grammar with $L=3$, $v=16$, $m=8$ and $m_{\mathrm{keep}}=2$ (\textbf{(a)}, \textbf{(d)}), $m_{\mathrm{keep}}=4$ (\textbf{(b)}, \textbf{(e)}), $m_{\mathrm{keep}}=6$ (\textbf{(c)}, \textbf{(f)}); mean $\pm$ std over $4$ seeds. \textbf{(a)} to \textbf{(c)} Deleted-rule generation rates at levels $\ell=1,2,3$ against fine-tuning step $t$; dotted line: $1-m_{\mathrm{keep}}/m$. \textbf{(d)} to \textbf{(f)} The same curves against $t\,m_{\mathrm{keep}}^{-\ell/2}$, with the same axes in the three panels. The probability of generating deleted rules starts decreasing at different timescales depending on $m_{\mathrm{keep}}$ and rule level $\ell$. Rescaling the fine-tuning time step by $m_{\mathrm{keep}}^{\ell/2}$ collapses the point where the curves start decreasing ((d) to (f)).}
    \label{fig:forbidden-L3}
\end{figure}

Figure~\ref{fig:app-level-collapse} shows the deleted-rule rates of
Figures~\ref{fig:forget-timescales}(a) and~\ref{fig:app-timescales}(a), (d)
against the rescaled step $t\,m_{\mathrm{keep}}^{-\ell/2}$ of \cref{eq:tft},
at full size and with the variation across seeds, for $m=6$, $8$ and $10$.
Figure~\ref{fig:forbidden-L3} shows the same comparison
for the grammar with $L=3$ at three values of $m_{\mathrm{keep}}$.

\FloatBarrier
\subsection{Linear probe}
\label{sec:app-probe}

As complementary evidence to the representation probe $q^R$ in the main text, we fit linear probes to the hidden representations \citep{davari2022probing} to test whether the symbol of a node can still be read from the representation when this node uses a deleted rule. The results are shown in Figures~\ref{fig:app-lr-probe}, \ref{fig:app-bounded} and~\ref{fig:app-wd}. The probe is a linear classifier that predicts the symbol of one latent node from the hidden representation at one position. We probe the three nodes of $q^R$, $\mu_{-2}^{\ell}$ with $\ell=1,2,3$. The probe is fitted on sequences $x$ sampled from $\Gft$, in which every node uses a kept rule, and it is tested on transformed sequences $Rx$: the rule of the probed node is replaced by a uniformly drawn deleted rule of the same symbol, the subtree of the node is generated again with kept rules, and all other tokens are unchanged. The probe therefore never sees a deleted rule during fitting, and its test accuracy is high only if the representations of the deleted rules are intact and represent the symbol in the same way as those of the kept rules. We call this test accuracy the linear probe accuracy.

The classifier is a multinomial logistic regression over the $v$ symbols, fitted with an $L_2$ penalty of coefficient $10^{-4}$ on its weights. Each coordinate of the representation is standardized with the mean and standard deviation of the fitting inputs. The test sequences $Rx$ are built from other samples $x$ than the fitting sequences, so that fitting and testing share no sequence; the numbers of sequences are given in Table~\ref{tab:app-settings}.

For each probed node, the representation is read at the last token of the node's subtree. The symbol of a node is determined by the tokens of its subtree, and the network at a given position has seen only the tokens up to that position. The last token of the subtree is therefore the first position at which the representation can contain the symbol of the node. We count token positions from the end of the sequence, as for the node indices: $x_{-1}$ is the last token and $x_{-2}$ the second to last. The subtree of $\mu_{-2}^{\ell}$ ends at the $(2^{\ell}+1)$th token from the end, that is, at $x_{-3}$, $x_{-5}$ and $x_{-9}$ for $\ell=1,2,3$. The probe $q^R$ is instead read at $x_{-2}$, the position that predicts the last token. We fit one probe at the input embedding and after each layer of the network, and report the largest test accuracy. Thus, if any representation of the deleted rules is still intact after any layer, it can be picked out.

\FloatBarrier
\subsection{Learning-rate dependence}
\label{sec:app-lr}

\begin{figure}[!htb]
    \centering
    \includegraphics{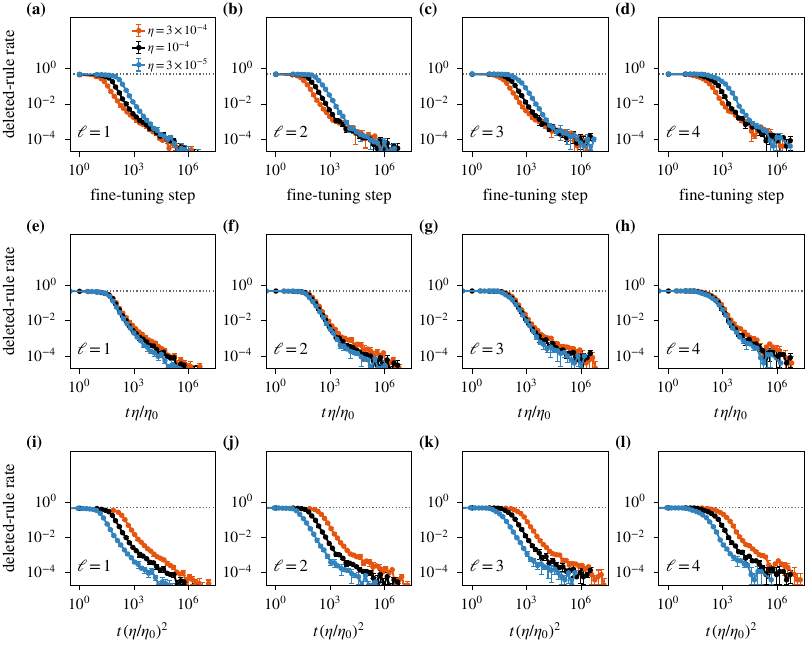}
    \caption{Deleted-rule rate at levels $\ell=1$ to $4$ (columns) in the restriction
    fine-tuning at $m=6$, $m_{\mathrm{keep}}=3$ with learning rates $\eta=3\times10^{-4}$,
    $10^{-4}$ and $3\times10^{-5}$, against the fine-tuning step $t$ (top row), $t\,\eta/\eta_0$
    (middle row) and $t\,(\eta/\eta_0)^2$ (bottom row), $\eta_0=10^{-4}$; dotted line: $1/2$;
    mean $\pm$ std over $3$ seeds.
    The middle row shows that the curves collapse on each other when plotted against $t\,\eta/\eta_0$, consistently with the hypothesis of Section~\ref{sec:forget-weights} that fine-tuning generation behavior is controlled by the average loss gradient.
    }
    \label{fig:app-lr-gen}
\end{figure}

\begin{figure}[!htb]
    \centering
    \includegraphics{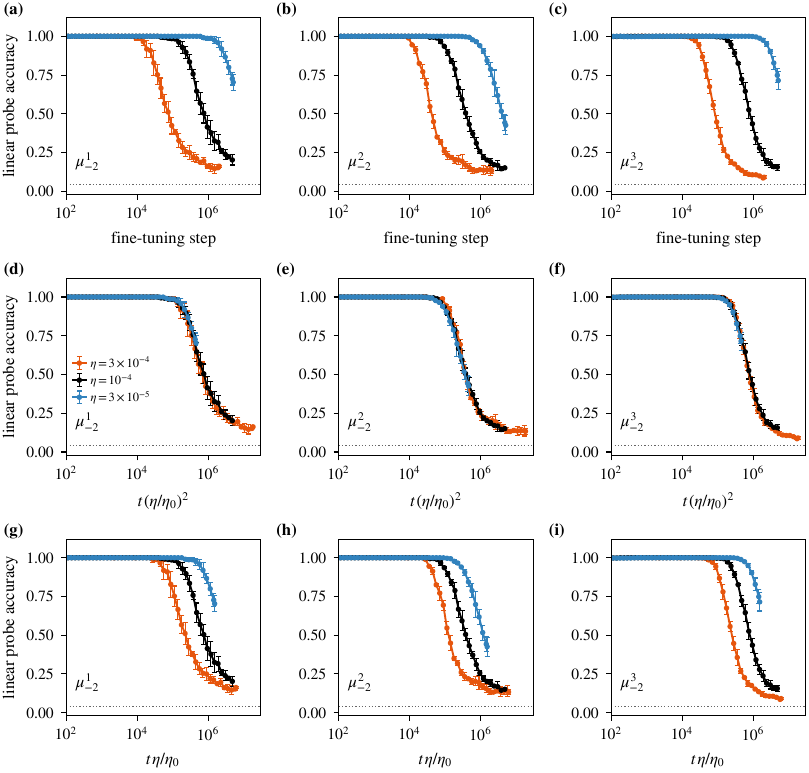}
    \caption{Linear probe accuracy (Appendix~\ref{sec:app-probe}) at $\mu_{-2}^1$,
    $\mu_{-2}^2$ and $\mu_{-2}^3$ (columns) for the learning rates of
    Figure~\ref{fig:app-lr-gen}, against the fine-tuning step $t$ (top row),
    $t\,(\eta/\eta_0)^2$ (middle row) and $t\,\eta/\eta_0$ (bottom row); dotted line: chance $1/v$; mean $\pm$ std over
    $3$ seeds.
    The middle row shows that the curves collapse on each other when plotted against $t\,(\eta/\eta_0)^2$, consistently with the hypothesis of Section~\ref{sec:forget-weights} that representational forgetting is controlled by the variance of the loss gradients.
    }
    \label{fig:app-lr-probe}
\end{figure}

Both the deleted-rule rate and the linear probe accuracy collapse when plotted against a rescaled time. Figure~\ref{fig:app-lr-gen} shows that, at every level, the deleted-rule rates at the three learning rates coincide under the scaling $t\,\eta/\eta_0$. The linear probe accuracies of the latent nodes at the different levels instead coincide under the scaling $t\,(\eta/\eta_0)^2$ (Figure~\ref{fig:app-lr-probe}). This difference supports the argument for two different regimes of behavioral suppression and representational drift in Section~\ref{sec:forget-weights}.
In each figure, the last row shows the other scaling under which the curves do not coincide.

\begin{figure}[!htb]
    \centering
    \includegraphics{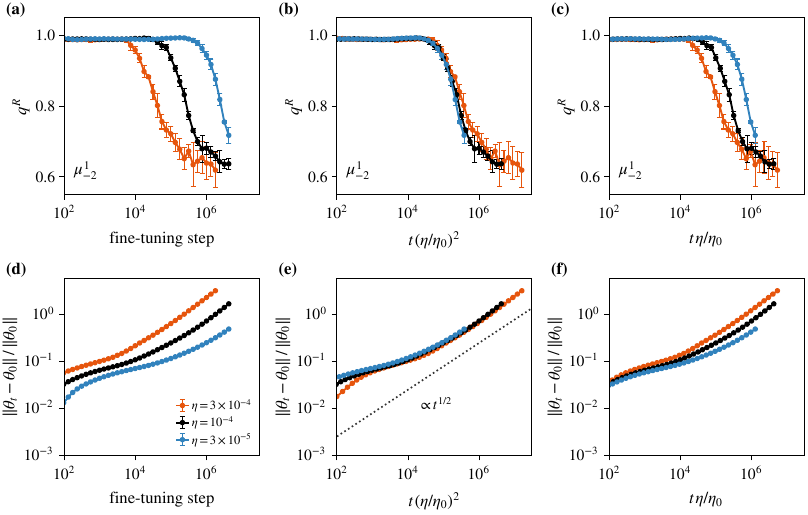}
    \caption{Representation probe $q^R$ at $\mu_{-2}^1$ (\textbf{(a)} to
    \textbf{(c)}) and distance from the pretrained parameters relative to their norm, $\|\theta_t-\theta_0\|/\|\theta_0\|$ (\textbf{(d)} to \textbf{(f)}), against the fine-tuning step $t$, for the learning rates $\eta$ of Figure~\ref{fig:app-lr-gen}. 
    Curves represent the mean $\pm$ std over $3$ seeds.
    Panels (a) and (d) show that larger $\eta$ corresponds to faster weight displacement, which in turn results in faster representational forgetting.
    Panels (b) and (e) show that rescaling the x-axis as $t\,(\eta/\eta_0)^2$ collapses both the representational forgetting and the weight displacements, which align to a single curve proportional to $t^{1/2}$ (dotted line). This is consistent with the discussion in Section~\ref{sec:forget-weights} that representational forgetting is driven by accumulated noise in the gradient updates.
    Panels (c) and (f) show that $t\,\eta/\eta_0$ is not the variable controlling this stage of dynamics (cf. \Cref{fig:app-lr-gen}).
    }
    \label{fig:app-lr-slow}
\end{figure}

Figure~\ref{fig:app-lr-slow} shows the same comparison for the two quantities of Figure~\ref{fig:forget-lr}(b) and (c), the representation probe $q^R$ at $\mu_{-2}^1$ and the distance of the parameters from their pretrained values. Both coincide under the scaling $t\,(\eta/\eta_0)^2$ and not under $t\,\eta/\eta_0$; for the distance, this is in the later stage, in which it increases approximately as $t^{1/2}$.

\FloatBarrier
\subsection{Bounded suppression}
\label{sec:app-bounded}

\begin{figure}[!htb]
    \centering
    \includegraphics{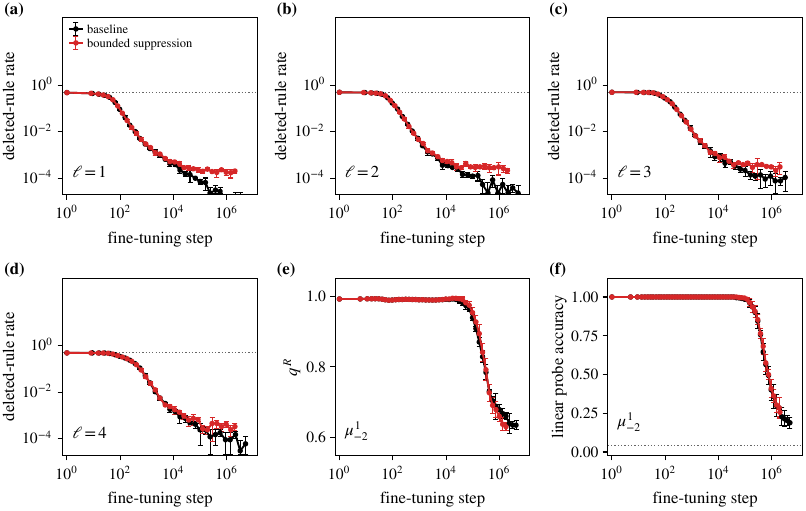}
    \caption{Restriction fine-tuning at $m=6$, $m_{\mathrm{keep}}=3$ for the baseline and for
    bounded suppression (Section~\ref{sec:forget-weights}), against the fine-tuning step $t$;
    mean $\pm$ std over $4$ seeds. \textbf{(a)} to \textbf{(d)} Deleted-rule rate at levels
    $\ell=1$ to $4$; dotted line: $1/2$. \textbf{(e)} $q^R$ at $\mu_{-2}^1$. \textbf{(f)} Linear probe
    accuracy at $\mu_{-2}^1$ (Appendix~\ref{sec:app-probe}); dotted line: chance $1/v$.
    With bounded suppression the deleted-rule rate stops decreasing at a small value, while $q^R$
    and the linear probe accuracy follow the baseline: the slow change in representations is not
    caused by the continued suppression of deleted rules.}
    \label{fig:app-bounded}
\end{figure}

Figure~\ref{fig:app-bounded} compares the baseline with the bounded suppression of Section~\ref{sec:forget-weights}, in which a token other than the target is dropped from the cross-entropy once its logit is more than $8$ nats below the largest logit of its position, so that the loss stops pushing tokens that are already suppressed. With bounded suppression, the deleted-rule rate stops decreasing at a small value at every level (Figure~\ref{fig:app-bounded}(a) to (d)), while $q^R$ at $\mu_{-2}^1$ follows the same curve as in the baseline (Figure~\ref{fig:app-bounded}(e)), and so does the linear probe accuracy at the same node (Figure~\ref{fig:app-bounded}(f), Appendix~\ref{sec:app-probe}).

\FloatBarrier
\subsection{Two regularizations in the fine-tuning}
\label{sec:app-wd}

Section~\ref{sec:forget-weights} relates the slow timescale to the drift of the parameters away from their pretrained values $\theta_0$. Here we change this drift with two explicit regularizations that act on it in opposite ways: one moves the parameters back toward $\theta_0$, the other moves them away from $\theta_0$, toward zero.

\paragraph{Setup.} The $L_2$ regularization toward the pretrained parameters moves each parameter a fraction $\eta\lambda$ of the way toward its pretrained value at every step, $\theta\leftarrow\theta+\eta\lambda\,(\theta_0-\theta)$. Decoupled weight decay \citep{loshchilov2019decoupled} is the same regularization toward zero, $\theta\leftarrow(1-\eta\lambda)\,\theta$. Acting alone, either regularization brings the parameters close to its target over a time of order $1/(\eta\lambda)$. Both are applied separately from the Adam update, to the matrices and embeddings but not to the layer-normalization gains. All other settings are unchanged.

\begin{figure}[!htb]
    \centering
    \includegraphics{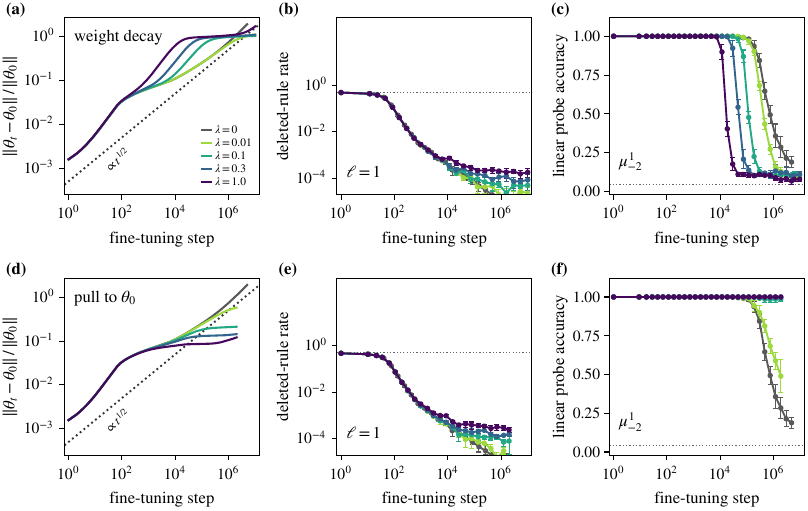}
    \caption{Restriction fine-tuning at $m=6$, $m_{\mathrm{keep}}=3$ with decoupled weight decay of coefficient $\lambda$ (top row) or an $L_2$ regularization of coefficient $\lambda$ toward the pretrained parameters $\theta_0$ (bottom row), both applied separately from the Adam update; legend in (a); $\lambda=0$ is the same baseline in both rows. \textbf{(a), (d)} Distance from the pretrained parameters relative to their norm, $\|\theta_t-\theta_0\|/\|\theta_0\|$; dotted line: $\propto t^{1/2}$. \textbf{(b), (e)} Deleted-rule rate at $\ell=1$; dotted line: $1/2$. \textbf{(c), (f)} Linear probe accuracy at $\mu_{-2}^1$ (Appendix~\ref{sec:app-probe}); dotted line: chance $1/v$. Mean $\pm$ std over $4$ seeds.
    Both regularizations leave the time-scale for the decrease of the deleted-rule rate unchanged ((b), (e)). Weight decay moves the parameters away from $\theta_0$ over a time of order $1/(\eta\lambda)$, and the linear probe accuracy falls earlier for larger $\lambda$ ((a), (c)). The regularization toward $\theta_0$ stops the increase of the distance and, when strong enough, keeps the accuracy close to $1$ ((d), (f)).}
    \label{fig:app-wd}
\end{figure}

\paragraph{The behavioral suppression is unchanged.} The suppression is completed long before a time of order $1/(\eta\lambda)$. Over this interval the deleted-rule rate and the distance from $\theta_0$ follow the same curves for every $\lambda$, under both regularizations (Figure~\ref{fig:app-wd}(a), (b), (d) and (e)). Afterwards the deleted-rule rate stays low, with only a slightly higher residual for larger $\lambda$, including under the strongest regularization toward $\theta_0$. The behavioral suppression thus requires only a small displacement from $\theta_0$, which the fine-tuning gradient maintains against the regularization.

\paragraph{Regularization toward $\theta_0$ delays or prevents the loss of the representations.} This regularization counteracts the drift: the distance from $\theta_0$ stops increasing, at a value that is smaller for larger $\lambda$ (Figure~\ref{fig:app-wd}(d)). When this value is small enough, the linear probe accuracy stays close to $1$ over the whole run, so the representations of the deleted rules are retained. Under the weakest regularization the accuracy still decreases, but later than in the baseline (Figure~\ref{fig:app-wd}(f)).

\paragraph{Weight decay accelerates the loss of the representations.} Weight decay adds a systematic motion of the parameters away from $\theta_0$, toward zero, over a time of order $1/(\eta\lambda)$ (Figure~\ref{fig:app-wd}(a)). When this time is shorter than the slow timescale of the baseline, the linear probe accuracy decreases after a time of order $1/(\eta\lambda)$, earlier for larger $\lambda$. When it is longer, the noise drift comes first, and the accuracy decreases only slightly earlier than without decay (Figure~\ref{fig:app-wd}(c)).

\FloatBarrier
\subsection{Recovering the deleted rules}
\label{sec:app-relearn}

We test whether a fine-tuned network generates the deleted rules again when it is trained again on $\Gpre$, and how the time that this takes depends on the duration of the fine-tuning.

\paragraph{Setup.} We take the network at fine-tuning step $t$, for checkpoints taken throughout the fine-tuning, reset the Adam moments, and train it on freshly sampled $\Gpre$ sequences with the fine-tuning settings. During this training we measure the deleted-rule rate at each level with the generation probe. Under $\Gpre$ this rate is $1/2$. The relearning time of level $\ell$ is the number of training steps until its deleted-rule rate first reaches $0.45$. As a reference, we train the same architecture on $\Gpre$ from random initialization with the same settings, which repeats the pretraining, and measure the same times.

\begin{figure}[!htb]
    \centering
    \includegraphics{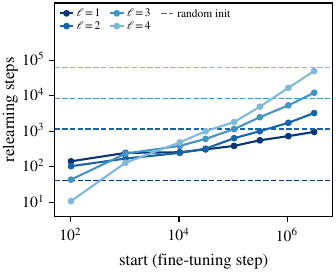}
    \caption{Retraining on $\Gpre$ from the restriction fine-tuning at step $t$ ($m=6$,
    $m_{\mathrm{keep}}=3$): steps until the deleted-rule rate of level $\ell$ is back above $0.45$
    ($1/2$ under $\Gpre$), against $t$; dashed lines: the same network trained on $\Gpre$ from
    random initialization, same color per level. Seed $0$.}
    \label{fig:app-relearn}
\end{figure}

\paragraph{Dependence on the duration of the fine-tuning.} At every level, the relearning time increases with the fine-tuning step $t$ from which the relearning starts (Figure~\ref{fig:app-relearn}). This increase begins while the linear probe accuracy is still close to $1$ (Figure~\ref{fig:app-lr-probe}), so the loss of the representations is not its only cause. Over the same interval, the loss on the deleted grammar keeps increasing (Figure~\ref{fig:forget-timescales}(c)): the fine-tuning makes the deleted rules less and less probable, and more steps are needed to undo this suppression.

\paragraph{Comparison with random initialization.} From the checkpoints taken after the fast suppression and before the slow timescale, the levels $\ell=2$ to $4$ are relearned faster than they are learned from random initialization (dashed lines in Figure~\ref{fig:app-relearn}), by a larger factor at higher levels. This advantage decreases for later starts. From the latest start, at which the representations of the deleted rules are lost (Figure~\ref{fig:app-lr-probe}), only the root is still relearned faster than from random initialization, and the levels $\ell=1$ to $3$ are relearned more slowly. The bottom level $\ell=1$ differs from the others at every start: it is learned within very few steps from random initialization, and relearning it from a fine-tuned network takes longer, because the suppression has to be undone.

In summary, between the two timescales the suppressed rules at the levels $\ell=2$ to $4$ are relearned much faster than they are learned from random initialization, and after the slow timescale this advantage is lost.

\FloatBarrier
\subsection{Continuations after deleted-rule context}
\label{sec:app-prompt}

We test whether a context that contains deleted rules leads the fine-tuned network to generate deleted rules again. This is the analogue in the RHM of prompting a fine-tuned model with text that it no longer generates.

\paragraph{Setup.} We sample a sequence from $\Gpre$ and give its first half $x_1,\dots,x_8$ to the fine-tuned network as context. Under $\Gpre$, each of the seven latent nodes of the first half uses a deleted rule with probability $1-m_{\mathrm{keep}}/m=1/2$. The network then generates the second half $x_9,\dots,x_{16}$ by sampling at temperature $1$. We parse the whole sequence with $\Gpre$ and assign each node of the second half to one of three cases: it uses a kept rule, it uses a deleted rule, or it is invalid, which means that its children are not a rule of $\Gpre$ or that a node below it is invalid. We use $8192$ first halves per checkpoint and seed.

We assign the root to the same three cases. The rule at the root joins the two halves: its left child is the top node of the first half, whose symbol $a$ is fixed by the context, and its right child is the top node of the generated half. A valid root therefore requires a second half whose top symbol forms a rule of $\Gpre$ with $a$. The root thus tests whether the network still identifies the symbol $a$ from a context written partly with deleted rules.

\begin{figure}[!htb]
    \centering
    \includegraphics{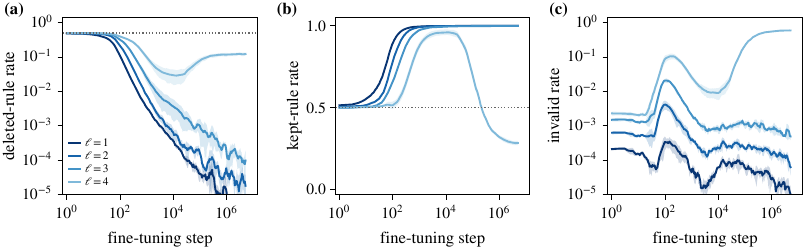}
    \caption{Free continuation of the second half after a first half drawn from $\Gpre$,
    in which each node uses a deleted rule with probability $1/2$ ($m=6$, $m_{\mathrm{keep}}=3$).
    Rate of deleted rules \textbf{(a)}, of kept rules \textbf{(b)} and of invalid nodes
    \textbf{(c)} at each level of the parsed continuation: $\ell=1,2,3$ inside the second half,
    $\ell=4$ the rule at the root, which joins the two halves. A node is invalid when its children
    are not a rule of $\Gpre$ or a node below it is invalid. Dotted lines: $1/2$, the value under
    $\Gpre$ in (a) and (b). Rates pooled over five consecutive checkpoints; mean $\pm$ std over
    $4$ seeds.}
    \label{fig:app-prompt}
\end{figure}

\paragraph{Inside the generated half.} At the levels $\ell=1,2,3$, once generation is suppressed, nearly every node of the generated half uses a kept rule (Figure~\ref{fig:app-prompt}(b)), and the deleted-rule rate keeps decreasing for the rest of the fine-tuning (Figure~\ref{fig:app-prompt}(a)). Deleted rules in the context therefore do not lead the network to generate deleted rules.

\paragraph{At the root.} After the fast suppression, the large majority of the roots use a kept rule (Figure~\ref{fig:app-prompt}(b), $\ell=4$). The network identifies the symbol $a$ of the first half, including when it is written with deleted rules, and continues the sequence with a kept rule. There are two exceptions. For a small fraction of first halves, no kept rule at the root has $a$ as its left child, so the sequence cannot be completed with a kept rule, and the network generates a deleted rule or an invalid root. For the other first halves, the root uses a deleted rule in a few percent of the generated sequences.

Much later in the fine-tuning, the fraction of invalid roots increases strongly (Figure~\ref{fig:app-prompt}(c)), over the interval in which the deleted-grammar-in-context loss increases (Figure~\ref{fig:forget-timescales}(c)). Invalid roots occur when the first half contains deleted rules, and are nearly absent when it contains none. The network thus no longer identifies the symbol of a first half written with deleted rules, and the second half that it generates no longer forms a rule with $a$. The deleted-rule rate at the root increases over the same interval (Figure~\ref{fig:app-prompt}(a)). This is not a return of the deleted rules: a second half generated without regard to $a$ gives an invalid root in most cases, and by chance a deleted rule in some.

\FloatBarrier
\subsection{Fine-tuning after resampling one rule book}
\label{sec:app-resample}

We redraw the rule book at one level $\ell$ independently from the
same RHM ensemble, leaving the other books unchanged. Some tuples or
rules can therefore recur. Unlike the restriction in
Section~\ref{sec:forget-setup}, this intervention introduces new
parent--child associations. We consider $\ell=1,\dots,4$ at $m=6$
and fine-tune for $1.25\times10^6$ steps, with four seeds per level
and the optimization settings of Section~\ref{sec:forget-setup}.
Here $\Gft$ denotes the resampled grammar. We evaluate the loss and
plateau references defined in Section~\ref{sec:metric} and the
symbol-replacement probe $q^S$ of Appendix~\ref{sec:app-pretrain},
using this grammar for both the original
inputs and the transformed subtrees. The plateau references are
averaged over $10^5$ held-out sequences per seed.

\begin{figure}[!htb]
    \centering
    \includegraphics{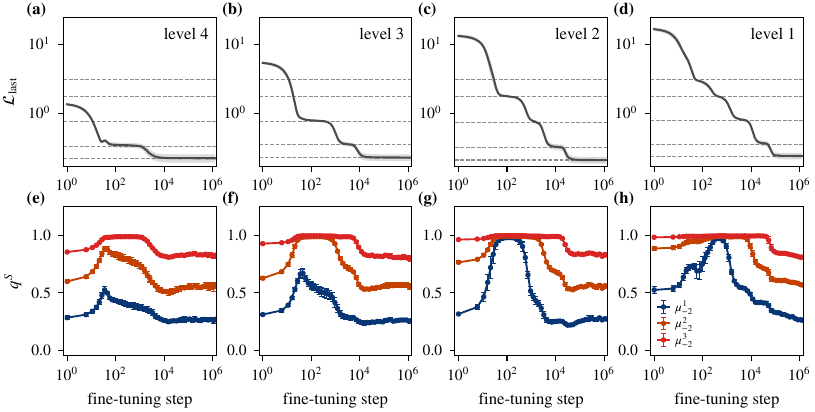}
    \caption{Fine-tuning after resampling one rule book at $m=6$;
    columns run from the root book (level $4$) to the bottom book
    (level $1$). Curves and uncertainties show mean $\pm$ std over
    $4$ seeds. \textbf{(a)} to \textbf{(d)} $\Llast$ on sequences
    from the resampled grammar $\Gft$; gray dashed lines: $\ln v$
    and the seed-averaged plateau losses
    $\mathcal L_1,\dots,\mathcal L_4$ of $\Gft$.
    \textbf{(e)} to \textbf{(h)} $q^S$ at
    $\mu_{-2}^1$, $\mu_{-2}^2$ and $\mu_{-2}^3$, using $\Gft$.
    All panels share the fine-tuning-step axis.}
    \label{fig:app-resample}
\end{figure}

In Figure~\ref{fig:app-resample}(a) to (d), after the initial
adjustment, the loss approaches the reference
$\mathcal L_{\ell-1}$ associated with the resampled level:
$\mathcal L_3$ for the root book, $\mathcal L_2$ for level $3$,
$\mathcal L_1$ for level $2$, and approximately the random-guess
loss for the bottom book. The unchanged books below the edit retain
their conditional expansions, allowing prediction from local
structure. Their symbol marginals can nevertheless change, so
resampling need not preserve the old local prediction probabilities.

For the probed nodes $\mu_{-2}^{\ell'}$ with $\ell'\geq\ell-1$,
$q^S$ rises toward $1$ (Figure~\ref{fig:app-resample}(e) to (h)),
indicating reduced dependence on these
latents of the new grammar. Lower nodes show smaller, transient
changes. Since the probe uses $\Gft$, it measures adaptation to
the new associations and does not establish complete loss of the
old representations.

The subsequent loss decreases pass through successive plateau
references, accompanied by decreases in the corresponding probes.
Resampling a lower book requires relearning more levels and takes
longer. This relearning is much slower than the initial adjustment,
whose duration also varies with the resampled level.

\FloatBarrier

\section{Generation fine-tuning and correlations}
\label{app:theory}

\subsection{Fine-tuning with fixed features}
\label{app:toy-model}

Consider predicting a token $y\in[v]$ from a context $x$
in an RHM sequence. Let $h$ denote the hidden latent at the
lowest common ancestor connecting the context to the target.

We consider an idealized $d$-dimensional fixed encoder learned during pretraining which recovers this latent from the context:
\[
\phi(x)=\phi_h,
\qquad
\frac{\phi_h^\top\phi_{h'}}{d}
=\kappa\,\delta_{hh'}.
\]
The predictor is a trainable linear readout,
\[
f_W(x)=W\phi(x),
\qquad
p_W(\cdot\mid x)=\operatorname{softmax}(f_W(x)),
\]
with $W\in\mathbb R^{v\times d}$.
Its predictions depend on the context only through the
encoded ancestor $h$. Training minimizes cross-entropy,
with $\phi$ held fixed.

\paragraph{Pretraining.}
We initialize the readout at pretrained weights $W^*$ that
approximately fit the pretraining distribution:
\[
p_h^*
:=
\operatorname{softmax}(W^*\phi_h)
\simeq p_{\mathrm{pre}}(\cdot\mid h).
\]

\paragraph{Fine-tuning.}
Following the restriction procedure in \Cref{sec:forget-setup},
only $m_{\mathrm{keep}}<m$ production rules are retained per
restricted parent. This changes the ancestor frequencies
and conditional targets to
\[
\pi_h=p_{\mathrm{ft}}(h),
\qquad
q_h=p_{\mathrm{ft}}(\cdot\mid h).
\]
Writing
\[
p_h=\operatorname{softmax}(W\phi_h),
\]
the population fine-tuning loss becomes
\[
\mathcal L(W)
=
-\mathbb E_{(x,y)\sim p_{\mathrm{ft}}}
\left[\log p_W(y\mid x)\right]
=
-\sum_h\pi_h\sum_y q_h(y)\log p_h(y).
\]

\paragraph{Initial readout update.}
The first population gradient update with learning rate $\eta$ reads
\[
W_1 - W_0
=
\eta\sum_h\pi_h(q_h-p_h)\phi_h^\top.
\]
At the pretrained initialization,

\[
W_1-W^*
=
\eta\sum_h\pi_h(q_h-p_h^*)\phi_h^\top.
\]
By orthogonality of the features, the corresponding logit
change for each ancestor is exactly
\begin{equation}
f_{1,h}-f_h^*
=
(W_1-W^*)\phi_h
=
\eta d\kappa\pi_h(q_h-p_h^*).
\label{eq:toy-first-logit-update}
\end{equation}

The ancestor marginals $\pi_h$ are random quantities of
typical size $1/v$. Approximating $\pi_h\simeq 1/v$, we get
\[
f_{1,h}-f_h^*
\simeq
\frac{\eta d\kappa}{v}
\left[
p_{\mathrm{ft}}(\cdot\mid h)
-p_{\mathrm{pre}}(\cdot\mid h)
\right],
\]
showing that the initial readout update is proportional to the change in the ancestor-conditioned target distribution.

\subsection{Computation of the correlation difference}
\label{app:conditional-change}

We estimate the RMS change in the conditional prediction,
\begin{equation}
\delta_\ell^2
:=\mathbb E\!\left[
\bigl(p_{\mathrm{ft}}^{(\ell)}(y\mid h)
-p_{\mathrm{pre}}^{(\ell)}(y\mid h)\bigr)^2
\right],
\label{eq:corr-difference-definition}
\end{equation}
where $\mathbb E$ averages over RHM rules.
As in the main text (\Cref{sec:theory}), $h$ denotes the context node and $\ell$ the depth of its lowest common ancestor with the target. 
For simplicity, we consider the case of a binary RHM tree ($s=2$) and retain $m_{\mathrm{keep}}$ of the original $m$ rules per parent at every relevant level. 
We use a balanced, large-vocabulary approximation: node marginals have typical size $1/v$, and their fluctuations are neglected in the leading estimate.

\paragraph{Case $\ell=1$.}
For $\ell=1$, the context is a visible token $c$ and the target $y$ is its sibling. Let $\mathcal S_c$ be the set of originally allowed successors, and let
$\mathcal S_c^{\mathrm{keep}}\subset\mathcal S_c$ be the retained set. Unambiguity ensures that each pair $(c,y)$ occurs in at most one production. Under uniform parent frequencies, the conditional is therefore uniform on the corresponding set of successors.
There are on average $m$ such successors before restriction and $m_{\mathrm{keep}}$ afterward. Replacing the counts by these values gives
\[
p_{\mathrm{pre}}(y\mid c)\simeq
\frac{\mathbf 1_{\{y\in\mathcal S_c\}}}{m},
\qquad
p_{\mathrm{ft}}(y\mid c)\simeq
\frac{\mathbf 1_{\{y\in\mathcal S_c^{\mathrm{keep}}\}}}
{m_{\mathrm{keep}}}.
\]
Thus a retained successor changes by $1/m_{\mathrm{keep}}-1/m$, whereas a deleted successor changes by $-1/m$. 
Summing the squared changes gives
\begin{align}
\sum_y\bigl[p_{\mathrm{ft}}(y\mid c)
-p_{\mathrm{pre}}(y\mid c)\bigr]^2
&\simeq
m_{\mathrm{keep}}
\left(\frac1{m_{\mathrm{keep}}}-\frac1m\right)^2
+\frac{m-m_{\mathrm{keep}}}{m^2} = 
\frac1{m_{\mathrm{keep}}}-\frac1m.
\label{eq:corr-one-level-count}
\end{align}
Dividing by $v$ yields the mean squared change per output token:
\begin{equation}
\delta_1^2\simeq
\frac1v\left(\frac1{m_{\mathrm{keep}}}-\frac1m\right),
\qquad
\delta_1\simeq
\frac{\sqrt{1-m_{\mathrm{keep}}/m}}
{\sqrt{v\,m_{\mathrm{keep}}}}.
\label{eq:corr-base-variance}
\end{equation}

\paragraph{General depth $\ell$.}
For $\ell>1$, the context identifies a latent $h$ on one side of the LCA, and the target descends from its other child through $\ell-1$ further levels. 
There are therefore approximately $m^\ell$
predictive paths before restriction and $m_{\mathrm{keep}}^\ell$
afterward, where each path counts sequences of rule choices.

Let $\mathcal P_{\mathrm{pre}}$ and
$\mathcal P_{\mathrm{ft}}\subset\mathcal P_{\mathrm{pre}}$ denote
these path sets. Let $I_\alpha(y)$ be one if path $\alpha$ ends
at $y$ and zero otherwise.
This indicator has disorder mean $1/v$ and variance
$(1/v)(1-1/v)\simeq1/v$. Suppressing the fixed arguments $(y\mid h)$,
the conditional probabilities are approximated by
\[
p_{\mathrm{pre}}^{(\ell)}\simeq
\frac1{m^\ell}\sum_{\alpha\in\mathcal P_{\mathrm{pre}}}I_\alpha,
\qquad
p_{\mathrm{ft}}^{(\ell)}\simeq
\frac1{m_{\mathrm{keep}}^\ell}
\sum_{\alpha\in\mathcal P_{\mathrm{ft}}}I_\alpha.
\]
Neglecting correlations between distinct paths, each additional
level reduces the variance by a factor $1/m$ or
$1/m_{\mathrm{keep}}$, respectively:
\begin{equation}
\operatorname{Var}\!\left(p_{\mathrm{pre}}^{(\ell)}\right)
\simeq\frac1{v m^\ell},
\qquad
\operatorname{Var}\!\left(p_{\mathrm{ft}}^{(\ell)}\right)
\simeq\frac1{v m_{\mathrm{keep}}^\ell}.
\label{eq:corr-single-grammar-variance}
\end{equation}
The two averages are not independent: all retained paths also
contribute to the pretrained prediction. Keeping their shared
contribution gives
\begin{equation}
\operatorname{Cov}\!\left(
p_{\mathrm{ft}}^{(\ell)},p_{\mathrm{pre}}^{(\ell)}
\right)
\simeq
\frac{|\mathcal P_{\mathrm{ft}}|}
{m_{\mathrm{keep}}^\ell m^\ell}\,\frac1v
\simeq\frac1{v m^\ell}.
\label{eq:corr-shared-rule-moment}
\end{equation}
Both conditional probabilities have disorder mean $1/v$.
Their mean squared difference therefore equals the sum of their
variances minus twice this covariance, giving
\begin{equation}
\boxed{
\delta_\ell^2\simeq\frac1v
\left(\frac1{m_{\mathrm{keep}}^\ell}-\frac1{m^\ell}\right),
\qquad
\delta_\ell\simeq
\frac{\sqrt{1-(m_{\mathrm{keep}}/m)^\ell}}
{\sqrt v\,m_{\mathrm{keep}}^{\ell/2}}.
}
\label{eq:corr-difference-leading}
\end{equation}
This recovers the one-level calculation and vanishes when no rules
are deleted. For a fixed retained fraction, the leading dependence on $m_{\mathrm{keep}}$ is $m_{\mathrm{keep}}^{-\ell/2}$.
Using $C_\nu^{(\ell)}(h,y)=p_\nu(h)
[p_\nu^{(\ell)}(y\mid h)-p_\nu(y)]$, the balanced approximation
relates the correlation difference to the conditional difference
by an additional factor $1/v$. Hence
\begin{equation}
\mathbb E\!\left[
\bigl(C_{\mathrm{ft}}^{(\ell)}-C_{\mathrm{pre}}^{(\ell)}\bigr)^2
\right]
\simeq\frac1{v^3}
\left(\frac1{m_{\mathrm{keep}}^\ell}-\frac1{m^\ell}\right).
\label{eq:corr-connected-difference}
\end{equation}

\subsection{Comparison of fine-tuning with non-linear vs linearized model}
\label{app:linearReadout}
\begin{figure}[h]
    \centering
    \includegraphics[width=1\linewidth]{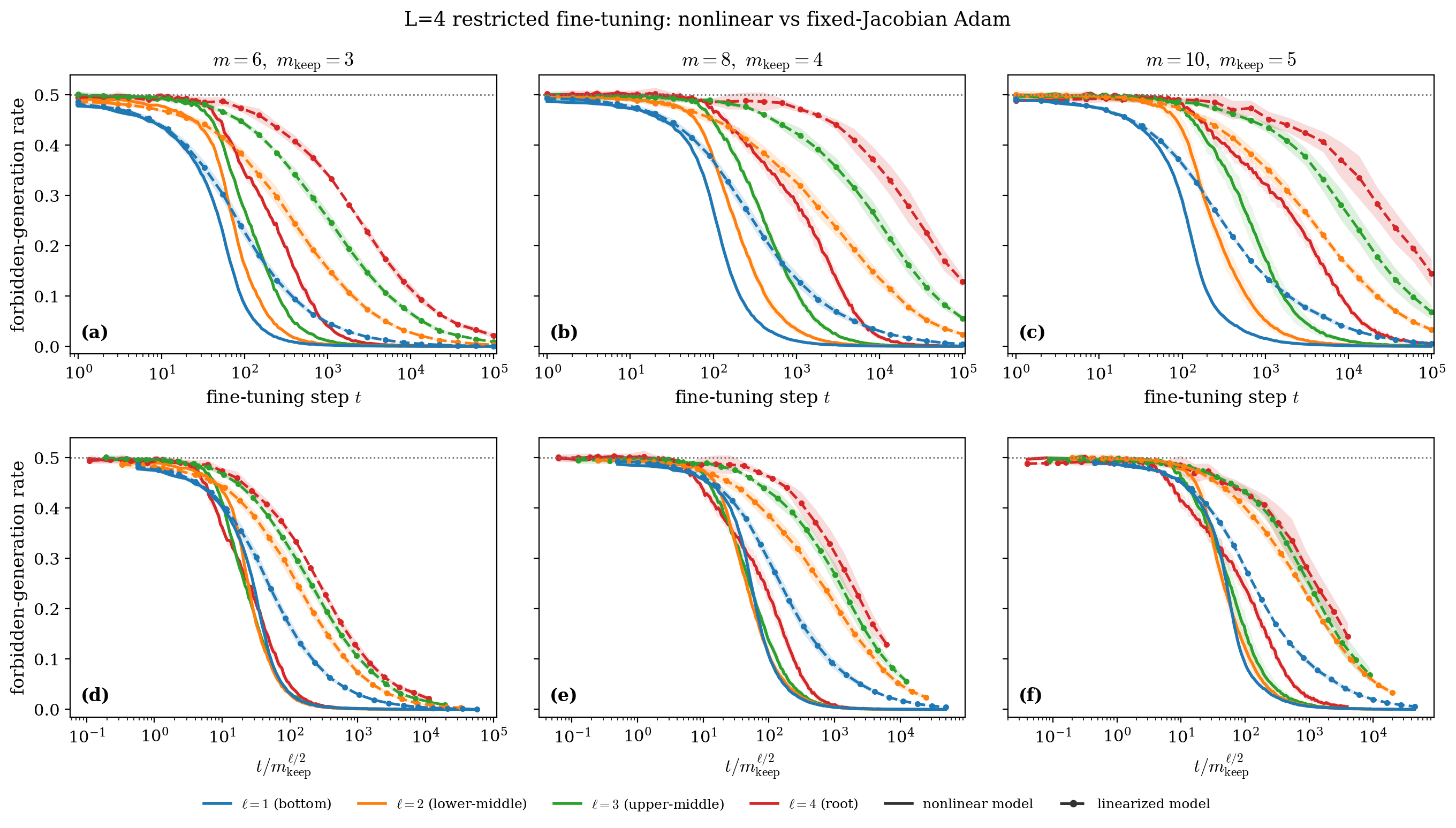}
    \caption{\textbf{Comparison of behavioral forgetting during fine-tuning for non-linear vs linearized model.} The linearized model \(f_{\mathrm{lin}}(x;\Delta\theta)=f_{\theta_0}(x)+J_{\theta_0}(x)\Delta\theta\) corresponds to a first order linearization around the pretrained weights $\theta_0$.
    Panels (a), (b), (c) show that only the adaptation timescale to forbidden rules at the bottom level is effectively described by the linearized model (i.e., the blue curves start to decay at the same time), while for higher hierarchical levels the linearized model is slower w.r.t. the non-linear one. Panels (d), (e), (f) show that the scaling $t^{(ft)}_{\ell}\sim m^{\ell/2}_{\mathrm{keep}}$ works across all levels only for the non-linear model (solid lines), while it captures some of the linear behavior only for increasing values of $m$ and $m_{\mathrm{keep}}$ (panel (f)).
    }
    \label{fig:liner_nonliner}
\end{figure}

To isolate the role of feature evolution during fine-tuning, we compare the full nonlinear model with its first-order linearization around the pretrained parameters \(\theta_0\). In the linearized experiment, the logits are approximated as \(f_{\mathrm{lin}}(x;\Delta\theta)=f_{\theta_0}(x)+J_{\theta_0}(x)\Delta\theta\), with the Jacobian \(J_{\theta_0}\) held fixed while \(\Delta\theta\) is optimized using Adam on the same restricted fine-tuning distribution. By contrast, the nonlinear model updates \(\theta\) directly, allowing both its representations and Jacobian to evolve. 
In \Cref{fig:liner_nonliner}, we consider \(L=4\) grammars with \(m=6,8,10\), retain half of the rules at every hierarchy level, and compare the level-wise forbidden-generation rates over the first \(10^5\) fine-tuning steps, averaged across four seeds. 
We observe that only for the hierarchical level $\ell=1$ the forbidden generation rate starts decreasing at the same timescale, while for larger $\ell$ the decreasing timescale is longer for the linearized model. This shows that, despite the fixed synonymic representations, feature adaptation is important for changing the forbidden rate generation.

\FloatBarrier

\section{Additional results and experimental details on bilingual small language model experiment}
\label{app:realistic-bilingual}
\subsection{Pretraining and fine-tuning of bilingual small language model}
\paragraph{Translation.}
We considered English-French (EN-FR) and English-Turkish (EN-TR) pairs. We primarily built the training corpus from \textit{uonlp/CulturaX}~\citep{nguyen-etal-2024-culturax}, a widely used multilingual web text collection after cleaning and deduplication. Among 2.21M respective documents of English, French and Turkish from the CulturaX corpora, we randomly sample 450,000 documents per language. Each document is split into sentences. For simplicity, we assume new sentences when followed by line breaks, then terminal punctuation followed by whitespace and
an upper-case letter, digit or opening quote, with an abbreviation list. We keep a document if it consists of at least 3 sentences and at least half of its sentences survive the criteria (at least 60\% letters are among non-space characters, no URL or e-mail address, at most 256 tokens, and a token-unigram language classifier that agrees with the document language), resulting in 427,631 English, 419,171 French documents, 422,975 Turkish documents. 
Every kept sentence is machine-translated into the other language pair, from English to French, French to English, from English to Turkish and Turkish to English with the corresponding
\textit{Marian Transformer-big}~\citep{tiedemann-2020-tatoeba,tiedemann-thottingal-2020-opus}, sentence by sentence without cross-sentence context. After the translation, we apply the same criteria again to each sentence, which keeps $\sim$97\% of the sentences. 
\paragraph{Bilingual text generation for pretraining.}
After the translation of each document, we rebuild the same document by randomly sampling each sentence's language. In other words, the contents of the document are the same, but the bilingual document switches each sentence's language between English and French randomly.  Therefore, language is half native and half translated. We train our own byte-level BPE tokenizer using English and French corpora of CulturaX, with a vocabulary of 16,384 tokens that yields comparable counts for English and French (0.248 tokens/char for English and 0.259/char for French). For English-Turkish pair, the CulturaX corpora with the same vocabulary of 16,384 tokens was used which again yields comparable counts (0.254 tokens/char for English and 0.249 tokens/char for Turkish). The EN-FR bilingual document has median 548 tokens and 18 sentences. EN-TR bilingual document has median 486 tokens and 17 sentences. 

\paragraph{Fine-tuning data.}
For fine-tuning, we use an unseen fraction of the data from CulturaX. After the pretraining with bilingual documents, we fine-tune the model with monolingual documents, reflecting the fine-tuning with data generated from restricted grammar in RHM. 
\paragraph{Training setup.}
The model is a decoder-only nanoGPT architecture~\citep{nanogpt2022} with 8 pre-LayerNorm blocks, 8
heads, width 512, MLP width 2048 with GELU activation, rotary position embeddings, no biases, no dropout, tied input/output
embeddings and context length of 512. This architecture resulted in $\sim$34M parameters. We train with the next-token prediction objective with batch size 64, for $10^5$ steps, corresponding to 3.28B tokens. For optimization, we used AdamW with $\beta$ = (0.9, 0.95), weight decay of $\{0, 0.1\}$, peak learning rate $6\times10^{-4}$, 1,000 warm-up steps, cosine decay to $6\times10^{-5}$, gradient clipping at 1.0.
For fine-tuning, we take the final checkpoints from the pretraining and continue next-token prediction training for $10^5$ steps at batch size of 64, with the constant learning rate $6\times10^{-5}$ and same optimizer setting. We perform 3 seeds of pretraining and apply a single seed of English monolingual fine-tuning and French/Turkish monolingual fine-tuning. 

\subsection{Evaluation of bilingual pretrained--monolingual fine-tuned language models}

\paragraph{Free generation.} We give a single end-of-text token in-context and let the model autoregressively generate 160 tokens at single context in total $1024$ samples. For each passage, we sample 160 tokens with temperature $1.0$. We count each sentence by splitting at newline tokens. To classify each sentence's language, we use a naive Bayes classifier over the vocabularies~\citep{raschka2014naive}. We take 50M tokens from EN and FR/TR each from the CulturaX corpus and count unigram frequency of each token in each language's corpus to estimate a categorical distribution over the vocabulary. Then we sum log-likelihood ratio between FR/TR and EN over the tokens in a sentence. A positive value means the sentence is likely French/Turkish and vice versa. We test that held-out monolingual sentences show $\sim99\%$ accuracy with a 10-token span using the classifier. We report the generation rate of the removed language during fine-tuning on both 1) the first sentence from generation and 2) the entire generated sentences. 
\paragraph{Representational similarity.} We subsample $6k$ single sentences from $6k$ unseen test passages. For the shuffled incoherent sentences, we keep the first and the last words of the sentence but shuffle the order of the remaining words, to minimize the effect due to a difference in the last word. We take the hidden representation after the final layer norm at the last word token of the sentence, center the hidden representation coordinate-wise and compute the absolute cosine similarity of the representations from coherent and shuffled sentences in each language.    
\paragraph{Prediction loss.} We use $6k$ unseen test passages in each language that consist of more than 3 sentences. We feed in all preceding sentences of each passage as a context either in EN monolingual or FR/TR monolingual, but not the last target sentence. Then, we measure the cross-entropy loss on the first token of the target sentence either in EN or FR/TR. This metric is a reflection of the three test families of the prediction loss (\Cref{fig:rhm-schematic}(f)). Assuming we fine-tune with EN: \textit{kept rule} is equivalent to measuring loss on EN target given EN context, \textit{deleted rule} is equivalent to measuring loss on FR/TR target given FR/TR context, and \textit{deleted in-context} measures loss on EN target given FR/TR context. 

\paragraph{Context ablation.} We reuse the $6k$ targets and contexts of the prediction loss and build, for each, an \textit{uninformative} context: every context sentence is replaced by a sentence drawn at random from a different test passage, in the same language. Then, we measure the cross-entropy loss on the averaged target sentence in the kept language. We report $\mathcal{L}_{\mathrm{uninformative}}-\mathcal{L}_{\mathrm{informative}}$, the loss under the uninformative context minus the loss under the original one either in the kept or in the removed language. We measure the same metric in a model that was only trained from the kept language which serves as a baseline.

\paragraph{Recoverability.} To test whether the removed language is suppressed or erased, we train the fine-tuned models again on the bilingual pretraining task and measure how fast the removed language recovers. The relearning corpus is built from the passages used for monolingual fine-tuning, but not included in the pretraining. We train for $10^4$ steps with the fine-tuning hyperparameters (constant learning rate $6\times10^{-5}$, batch of $64$ sequences of $512$ tokens, weight decay $0.1$). We start from the monolingual fine-tuned model at steps $10^2$, $10^3$, $10^4$ and $10^5$, and from three baselines: the bilingual pretrained model, which gives the level to recover; a model pretrained on the kept language alone, which has nothing to recover; and a randomly initialized model (trained from scratch). Along relearning, we measure the loss on a target sentence in the removed language given its context in the removed language, on the same $6k$ targets, averaged over the tokens after the first one.

\subsection{Additional results on bilingual pretrained--monolingual fine-tuned language models}
\begin{figure}[!htbp]
    \centering
    \includegraphics[width=\linewidth]{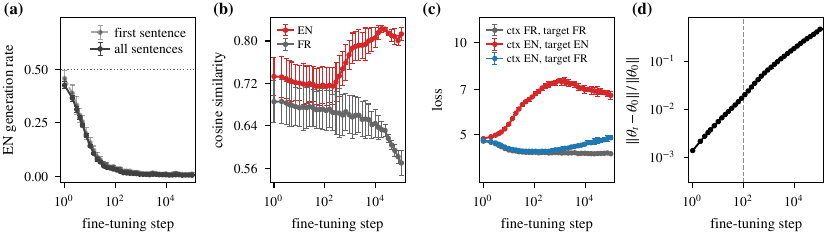}
    \caption{The measurements of Figure~\ref{fig:realistic} for
    French-only fine-tuning. \textbf{(a)} English generation rate.
    \textbf{(b)} Coherent--shuffled representation similarity for English
    (red) and French (gray). \textbf{(c)} Prediction loss for French
    targets with French context (gray), English targets with English
    context (red), and French targets with English context (blue).
    \textbf{(d)} Relative parameter distance, as in
    Figure~\ref{fig:realistic}(d).}
    \label{fig:app-realistic-fr}
\end{figure}

\begin{figure}[!htbp]
    \centering
    \includegraphics[width=\linewidth]{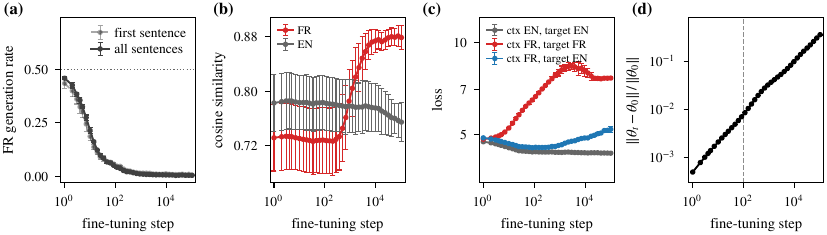}
    \caption{Bilingual pretraining and English-only fine-tuning without
    weight decay. Panels (a) to (d), colors and reference line as in
    Figure~\ref{fig:realistic}: French generation, coherent--shuffled
    representation similarity, prediction loss and relative parameter
    distance.}
    \label{fig:app-realistic-nowd-en}
\end{figure}

\begin{figure}[!htbp]
    \centering
    \includegraphics[width=\linewidth]{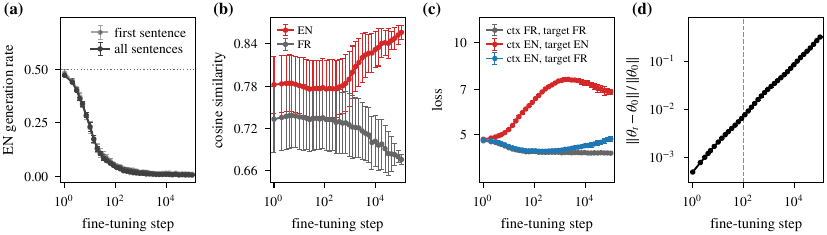}
    \caption{Bilingual pretraining and French-only fine-tuning without
    weight decay. Panels (a) to (d), colors and reference line as in
    Figure~\ref{fig:app-realistic-fr}: English generation,
    coherent--shuffled representation similarity, prediction loss and
    relative parameter distance.}
    \label{fig:app-realistic-nowd-fr}
\end{figure}

\begin{figure}[!htbp]
    \centering
    \includegraphics[width=\linewidth]{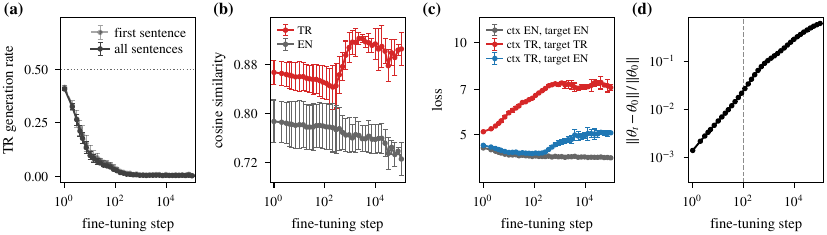}
    \caption{EN-TR bilingual pretraining and English-only fine-tuning. Panels (a) to (d), colors and reference line as in Figure~\ref{fig:realistic}: Turkish generation,
    coherent--shuffled representation similarity, prediction loss and
    relative parameter distance.}
    \label{fig:app-realistic-wd-entr-en}
\end{figure}

\begin{figure}[!htbp]
    \centering
    \includegraphics[width=\linewidth]{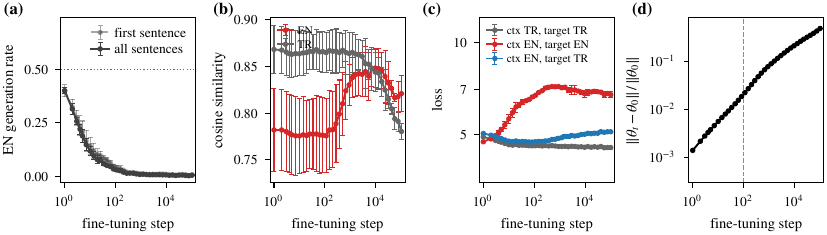}
    \caption{EN-TR bilingual pretraining and Turkish-only fine-tuning. Panels (a) to (d), colors and reference line as in
    Figure~\ref{fig:app-realistic-fr}: English generation,
    coherent--shuffled representation similarity, prediction loss and
    relative parameter distance.}
    \label{fig:app-realistic-wd-entr-tr}
\end{figure}

\begin{figure}[!htbp]
    \centering
    \includegraphics[width=\linewidth]{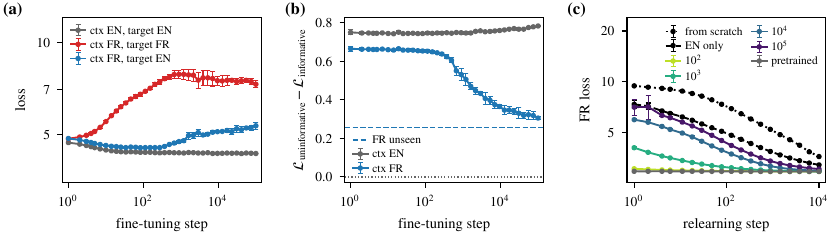}
    \caption{Bilingual pretraining and English-only fine-tuning additional control experiments. \textbf{(a)} The same three test families from \Cref{fig:realistic}(c) to recap the generation and representation forgetting timescales. \textbf{(b)} Context ablation experiment. At the onset of representational forgetting, removed language (FR) does not benefit from coherent, informative context. Before that, it benefits as much as the kept language (EN), and at the end of the fine-tuning it almost reaches the baseline level similar to the model which was never trained with French. \textbf{(c)} Recoverability experiment shows how much training is required to relearn the removed language after each fine-tuning. Before the onset of representational forgetting ($10^2$), the model almost immediately recovers to the level of the bilingual pretrained model (gray line). Relearning takes significantly more after the onset of representational forgetting ($10^3\sim$), almost similar to the model that was never trained with FR but only EN (black solid line). EN knowledge requires less relearning data than training from scratch indicating EN provides structural similarity to FR and enables the faster learning (black dashed line).}
    \label{fig:app-realistic-controls-en}
\end{figure}

\begin{figure}[!htbp]
    \centering
    \includegraphics[width=\linewidth]{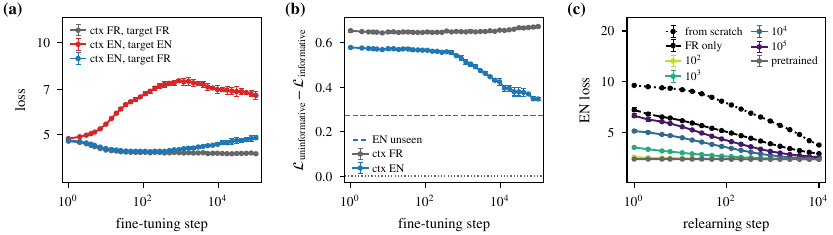}
    \caption{Bilingual pretraining and French-only fine-tuning additional control experiments. Panels (a)-(c) are equivalent to~\Cref{fig:app-realistic-controls-en}, only switching kept language (FR) and removed language (EN). }
    \label{fig:app-realistic-controls-fr}
\end{figure}

In the main text (\Cref{sec:realistic}), we only show the results from the EN-fine-tuned model. We report the FR/TR-fine-tuned model (\Cref{fig:app-realistic-fr}; \Cref{fig:app-realistic-wd-entr-en,fig:app-realistic-wd-entr-tr} for the English-Turkish pair) and also the models trained without weight decay (\Cref{fig:app-realistic-nowd-en,fig:app-realistic-nowd-fr}), which show a consistent trend. The control experiments are shown in \Cref{fig:app-realistic-controls-en,fig:app-realistic-controls-fr}.

\section{Additional results and experimental details on coding language model experiment}
\label{app:realistic-coding}
\subsection{Targeted formatting fine-tuning}
\paragraph{Fine-tuning data}
We use Qwen2.5-Coder 0.5B base model~\citep{hui2024qwen2}.
We build the fine-tuning corpus from \textit{codeparrot/github-code-clean}~\citep{CodeParrot}, a public collection of GitHub source files. We filter Python files and we rewrite two surface formatting conventions into one canonical form each, leaving the program unchanged: indentation with tabs or with 2, 3 or 8 spaces becomes 4 spaces; \texttt{\%} and \texttt{.format()} string formatting becomes f-strings. With this, for indentation formatting restriction 4-spaces becomes \textit{kept rule} and all others (tabs, 2, 3, or 8 spaces) become \textit{deleted rule}. For the string formatting restriction, f-strings is \textit{kept rule} and \texttt{\%} and \texttt{.format()} become \textit{deleted rule}. We use standard tools (\texttt{lib2to3}, \texttt{ruff}, \texttt{flynt}, \texttt{docstring\_parser}) which performs deterministic conversion between the formatting styles. We exclude a file if it does not parse or if any removed rule survives the conversion. This keeps 835,104 of 902,051 files (93\%), makes total 1.95B tokens. We report results for the two removed indentation formats (tabs, 2 spaces) and the two removed string-formatting formats. 
\paragraph{Model and training setup.}
We fine-tune all parameters of \textit{Qwen2.5-Coder-0.5B, 1.5B}~\citep{hui2024qwen2} base models, with either the indentation formatting restriction or string formatting restriction. We use AdamW ($\beta_1=0.9$, $\beta_2=0.95$, no weight decay) with learning rate $3\times10^{-5}$, held constant after a linear warm-up of 100 steps, batch size 32 sequences of 2,048 tokens, and gradient-norm clipping at 1. We train for 30,000 steps which roughly gives one pass over the fine-tuning dataset corpus.
\paragraph{Evaluation}
We follow the same evaluation metrics used for bilingual-transformer; (i) removed rule generation rate, (ii) cosine similarity between coherent / shuffled context with kept/removed rules and (iii) prediction loss after the context with kept/removed rule.
\subsection{Results on formatting fine-tuned large language models}

\begin{figure}[!htbp]
    \centering
    \includegraphics[width=\linewidth]{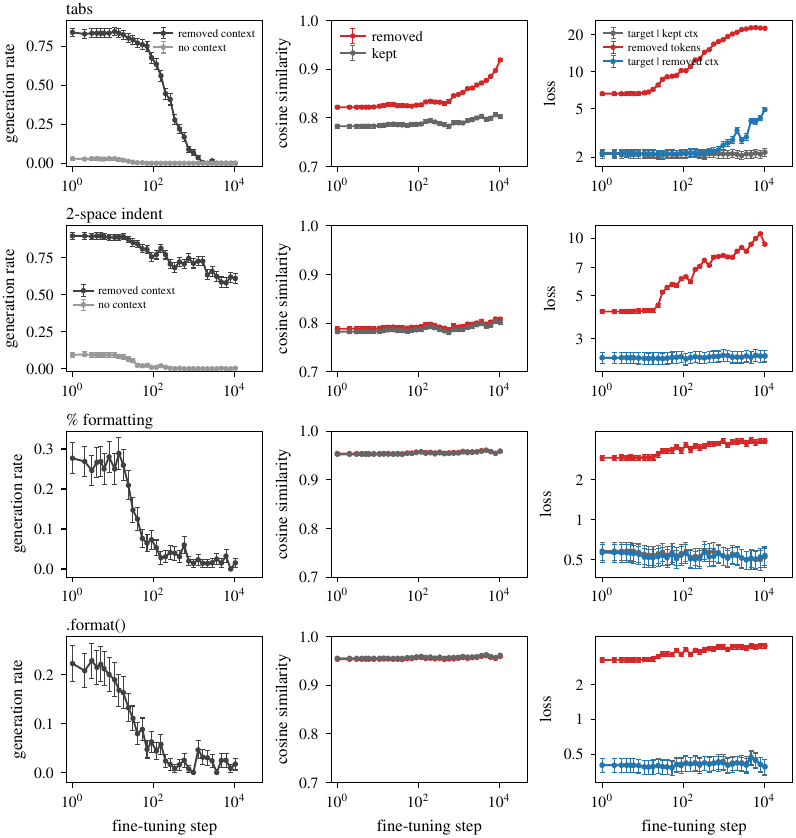}
    \caption{Targeted syntax fine-tuning on Qwen2.5-Coder-0.5B. Panels (a)-(d) are equivalent metrics to those of bilingual transformer experiments. The text on each panel indicates removed syntax.}
    \label{fig:app-realistic-coder-0p5B}
\end{figure}

\begin{figure}[!htbp]
    \centering
    \includegraphics[width=\linewidth]{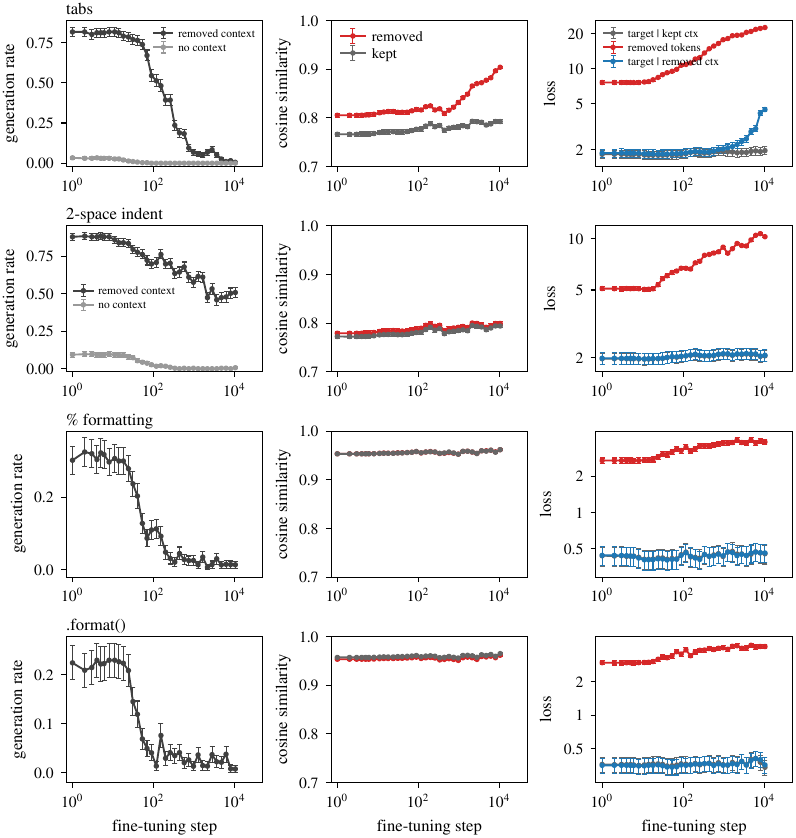}
    \caption{Targeted syntax fine-tuning on Qwen2.5-Coder-1.5B. Panels (a)-(d) are equivalent metrics to those of bilingual transformer experiments. The text on each panel indicates removed syntax.}
    \label{fig:app-realistic-coder-1p5B}
\end{figure}

\end{document}